\def\year{2024}\relax
\documentclass[letterpaper]{article} % DO NOT CHANGE THIS
\usepackage{aaai24}  % DO NOT CHANGE THIS
\usepackage{times}  % DO NOT CHANGE THIS
\usepackage{helvet}  % DO NOT CHANGE THIS
\usepackage{courier}  % DO NOT CHANGE THIS
\usepackage[hyphens]{url}  % DO NOT CHANGE THIS
\usepackage{graphicx} % DO NOT CHANGE THIS
\usepackage{natbib}  % DO NOT CHANGE THIS AND DO NOT ADD ANY OPTIONS TO IT
\usepackage{caption} % DO NOT CHANGE THIS AND DO NOT ADD ANY OPTIONS TO IT
\DeclareCaptionStyle{ruled}{labelfont=normalfont,labelsep=colon,strut=off} % DO NOT CHANGE THIS
\usepackage{algorithm, algorithmic}
\usepackage{amsmath, amssymb}
\usepackage{multirow, booktabs}
\usepackage{subcaption}
\usepackage{placeins}

\newtheorem{problem}{Problem}

\usepackage{newfloat}
\usepackage{listings}
\floatstyle{ruled}
\newfloat{listing}{tb}{lst}{}
\floatname{listing}{Listing}
\title{Robust Node Representation Learning via Graph Variational Diffusion Networks}
\author{Jun Zhuang, Mohammad Al Hasan}
\affiliations{
Boise State University,
Indiana University-Purdue University Indianapolis \\
junzhuang@boisestate.edu, alhasan@iupui.edu
}

\begin{document}

\maketitle

\begin{abstract}
Node representation learning by using Graph Neural Networks (GNNs) has been widely explored. However, in recent years, compelling evidence has revealed that GNN-based node representation learning can be substantially deteriorated by delicately-crafted perturbations in a graph structure. To learn robust node representation in the presence of perturbations, various works have been proposed to safeguard GNNs. Within these existing works, Bayesian label transition has been proven to be more effective, but this method is extensively reliant on a well-built prior distribution. The variational inference could address this limitation by sampling the latent node embedding from a Gaussian prior distribution. Besides, leveraging the Gaussian distribution (noise) in hidden layers is an appealing strategy to strengthen the robustness of GNNs. However, our experiments indicate that such a strategy can cause over-smoothing issues during node aggregation. In this work, we propose the Graph Variational Diffusion Network (GVDN), a new node encoder that effectively manipulates Gaussian noise to safeguard robustness on perturbed graphs while alleviating over-smoothing issues through two mechanisms: Gaussian diffusion and node embedding propagation. Thanks to these two mechanisms, our model can generate robust node embeddings for recovery. Specifically, we design a retraining mechanism using the generated node embedding to recover the performance of node classifications in the presence of perturbations. The experiments verify the effectiveness of our proposed model across six public datasets.
\end{abstract}

%\section{Introduction}
\section{Introduction}
\label{sec:intro}

Various web data can be represented as attributed graphs, such as citation networks~\cite{sen2008collective}, co-purchase networks~\cite{shchur2018pitfalls}, and online social networks (OSNs)~\cite{hamilton2017inductive}; the entities in these networks can be characterized as vertices, whereas the relationship between two entities can be depicted as a link.
As a powerful Artificial Intelligence (AI) technique for learning node representations in attributed graphs, Graph Neural Networks (GNNs)~\cite{bruna2013spectral, defferrard2016convolutional, kipf2016semi} are widely explored in numerous scenarios, such as user profiling in OSNs~\cite{chen2019semi, zhuang2022does}, community detection~\cite{breuer2020friend}, community-based question answering~\cite{fang2016community}, recommendation systems~\cite{ying2018graph}, etc.
A capstone of GNNs is the message-passing mechanism, which propagates the topology-based hidden representation by aggregating the node and features layer by layer.
In recent years, however, compelling evidence reveals that this mechanism may substantially deteriorate by delicately-crafted perturbations in a graph structure~\cite{sun2018adversarial, jin2020adversarial}, raising concerns about the reliability and robustness of GNNs.
For instance, in OSNs, non-malicious users may randomly connect with many other users for business promotion purposes~\cite{zhuang2022deperturbation}. On the other extreme, the message-passing mechanism may perform poorly on new OSN users who lack sufficient connections with other users (edges) or profile information (feature)~\cite{tam2020fiedler, ye2021sparse}. Third, adversarial attacks may significantly deteriorate the performance of GNNs with unnoticeable modification of graphs~\cite{zugner2018adversarial, dai2018adversarial}.

Many recent works have been proposed to learn robust node representation against the above-mentioned perturbations. We categorize these works from the perspective of processing stages as follows.
Topological denoising methods~\cite{wu2019adversarial, entezari2020all, rong2019dropedge} preprocess the poisoned graphs by eliminating suspicious edges before node aggregation in the pre-processing stage. However, these kinds of methods may fail to prune suspicious edges when the graph structure is sparse.
Mechanism design methods generally propose mechanisms, such as node aggregators~\cite{feng2020graph, jin2021node}, and regularization~\cite{chang2021not, liu2021elastic}, to strengthen the robustness of node classifications in the inter-processing stage. Nevertheless, these methods heavily rely on heuristic explorations of topological structures of graphs and may perform worse under severe perturbations.
Bayesian label transition methods~\cite{zhuang2022deperturbation, zhuang2022defending} tackle the perturbation issues in the post-processing stage by aligning the inferred labels to the original labels as closely as possible through MCMC-based sampling. The post-processing inference is proved to yield better accuracy on perturbed graphs, but this inference highly depends on the quality of predictions on the training data because MCMC methods indirectly sample the labels from the prior distribution, which is built on the training set~\cite{van2018simple}. Such a limitation could be addressed by variational inference~\cite{kingma2013auto} that samples from a robust prior distribution, e.g., Gaussian distribution, which is parameterized by mean $\mu$ and standard deviation $\sigma$.
Besides, leveraging the Gaussian distribution, a.k.a. Gaussian noise, to strengthen the robustness of GNNs is an appealing strategy~\cite{zhu2019robust, godwin2021simple, bo2021beyond}. To verify the effectiveness of this strategy, we conduct ablation experiments by adjusting the weight of Gaussian noise in the hidden layers of GNNs. The results suggest that simply increasing the percentage of Gaussian noise in the hidden layers tends to deteriorate the node classification accuracy. We argue that injecting Gaussian noise indiscriminately would cause over-smoothing issues in node aggregation. Based on this hypothesis, we face a core challenge: How to take advantage of Gaussian noise to enhance robustness against perturbations while simultaneously alleviating over-smoothing issues?

To embrace this challenge, we propose the Graph Variational Diffusion Network (GVDN), a novel node encoder that dynamically regulates the magnitude of noise and further safeguards robustness via two mechanisms: Gaussian diffusion and node embedding propagation. 
The first mechanism draws inspiration from variational diffusion models, which are commonly used in image synthesis~\cite{ho2020denoising}. The motivation for Gaussian diffusion is that the strategy of leveraging Gaussian noise may require less help from the noise as training progresses (the model becomes stronger). Gaussian diffusion linearly decreases the extent of Gaussian noise at each successive iteration to help train a robust model. Our experiments unearth that applying diffusion to $\mu$ and $\sigma$ before variational sampling via reparameterization can affirmatively help generalize the robustness of recovery on perturbed graphs.
The second mechanism is motivated by a recent study~\cite{zhuang2022robust}, which empirically verifies that label propagation can alleviate node vulnerability. Such propagation is based on the assumption of the graph homophily property~\cite{zhu2020beyond} that adjacent vertices tend to have similar class labels. Experiments indicate that propagating adjacent node embeddings of inaccurately predicted nodes, whose adjacent neighbors are selected by a topology-based sampler, during training can further strengthen the robustness.

Owing to the two mechanisms, our model can generate robust node embedding for downstream tasks. One benefit is that in real-world web applications, node embedding can be generated by an offline-trained GNN and further used for online downstream tasks, such as recommendation~\cite{sun2018recurrent} or search~\cite{xiong2017explicit}.
In this study, our downstream task is to recover the performance of node classifications on perturbed graphs; specifically, we design a retraining mechanism to recover the performance by retraining the GNN-based node classifier on perturbed graphs using the previously generated node embedding.
One advantage of our model over MCMC methods, such as GraphSS~\cite{zhuang2022defending}, is that our model doesn't require a well-built prior distribution from the training data. Besides, the experiments verify that our proposed retraining mechanism outperforms all competing methods in accuracy on perturbed graphs across all datasets.
Overall, our {\bf contributions} can be summarized as follows:
\vspace{-0.1cm}
\begin{itemize}
  \item We propose a new node encoder, namely the Graph Variational Diffusion Network (GVDN), to generate node embedding for robust node representation learning. To the best of our knowledge, our work is the first model that leverages variational diffusion with node embedding propagation to safeguard robustness against perturbations while alleviating over-smoothing issues on GNNs.
  \item We design a retraining mechanism using the generated node embedding to recover the performance of node classifications on perturbed graphs. Experiments validate the effectiveness of our model across six public datasets.
\end{itemize}

%\section{Methodology}
\section{Methodology}
\label{sec:method}
In this section, we first briefly introduce the notations. Most importantly, we introduce our proposed model in detail from the perspectives of frameworks, optimization, training procedure, retraining mechanism, and time complexity.

%\subsection{Notations}
%\label{subsec:notations}
\noindent
\textbf{\large Notations.}
We denote an undirected attributed graph as $\mathcal{G}$, which consists of a set of vertices $\mathcal{V}$ = $\{ v_{1}, v_{2}, ..., v_{N} \}$ and edges $\mathcal{E} \subseteq \mathcal{V} \times \mathcal{V}$, where $N$ is the number of vertices in $\mathcal{G}$. The topological relationship among vertices can be represented by a symmetric adjacency matrix $\mathbf{A} \in \mathbb{R}^{N \times N}$. We preprocess the adjacency matrix as follows:
\begin{equation}
\mathbf{\hat{A}} = \mathbf{\tilde{D}}^{-\frac{1}{2}} \mathbf{\tilde{A}} \mathbf{\tilde{D}}^{-\frac{1}{2}},
\label{eqn:hat_a}
\end{equation}
where $\mathbf{\tilde{A}} = \mathbf{A} + I_{N}$, $\mathbf{\tilde{D}} = \mathbf{D} + I_{N}$. $I_{N}$ is an identity matrix. $\mathbf{D}_{i,i} = \sum_{j} \mathbf{A}_{i,j}$ is a diagonal degree matrix.

In an attributed graph, each vertex contains a feature vector. All feature vectors constitute a feature matrix $\mathbf{X} \in \mathbb{R}^{N \times d}$, where $d$ is the number of features for each vertex. Given $\mathbf{\hat{A}}$ and $\mathbf{X}$, a layer-wise message-passing mechanism of GNNs can be formulated as follows:
\begin{equation}
\textit{f}_{\mathbf{W}^{(l)}} \left(\mathbf{\hat{A}}, \mathbf{H}^{(l)} \right) = \mathbf{\hat{A}} \mathbf{H}^{(l)} \mathbf{W}^{(l)},
\label{eqn:mpm}
\end{equation}
where $\mathbf{H}^{(l)}$ is a node hidden representation in the $l$-th layer. The dimension of $\mathbf{H}^{(l)}$ in the input layer, middle layer, and output layer is the number of features $d$, hidden units $h$, and classes $K$, respectively. $\mathbf{H}^{(0)} = \mathbf{X}$. $\mathbf{W}^{(l)}$ is the weight matrix in the $l$-th layer.

In this work, we aim to develop a node encoder to generate robust node embedding, a.k.a. node hidden representation, for the downstream node classification task on perturbed graphs. In general node classification tasks, a GNN takes both $\mathbf{\hat{A}}$ and $\mathbf{X}$ as inputs and is trained with {\bf ground-truth labels} $\mathbf{Y} \in \mathbb{R}^{N \times 1}$. Besides, we denote the {\bf predicted labels} as $\mathbf{\hat{Y}} \in \mathbb{R}^{N \times 1}$ (pronounced as "Y hat") and the {\bf sampled labels} as $\mathbf{\bar{Y}} \in \mathbb{R}^{N \times 1}$ (pronounced as "Y bar").
We formally state the problem that we aim to address as follows:
\begin{problem}
Given a graph $\mathcal{G} = (\mathbf{\hat{A}}, \mathbf{X})$ and the ground-truth labels $\mathbf{Y}$, we train our proposed node encoder on unperturbed $\mathcal{G}_{train}$ to generate node embedding for recovering the node classification performance of the GNN on perturbed $\mathcal{G}_{test}$.
\end{problem}

\begin{figure}[h]
  \centering
  \includegraphics[width=\linewidth]{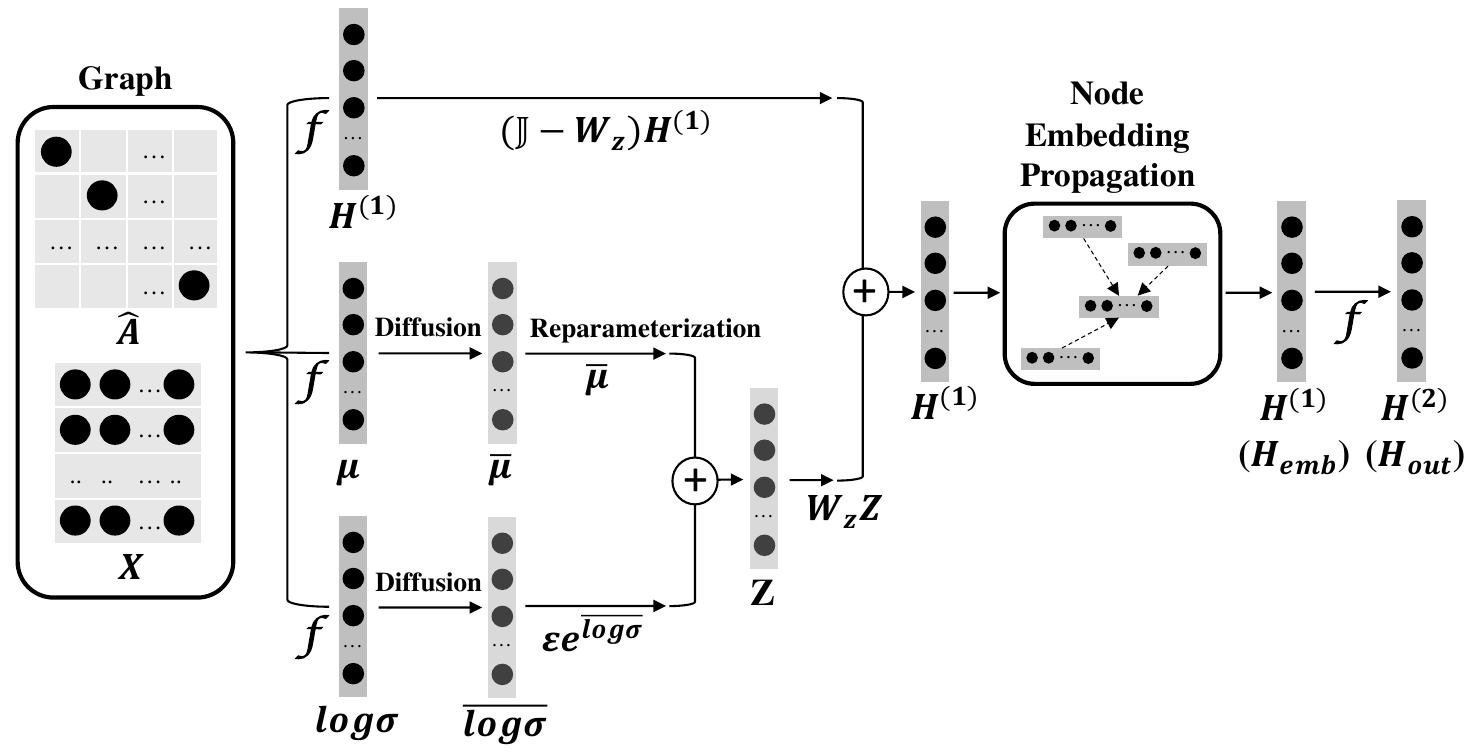}
  \caption{The overall framework of our proposed model. Given a preprocessed adjacency matrix $\mathbf{\hat{A}}$ and a feature matrix $\mathbf{X}$, we first generate the first-layer node hidden representation $\mathbf{H}^{(1)}$, mean matrix $\mathbf{\mu}$, and log standard deviation matrix $\mathbf{log\sigma}$ by node aggregation $f$ with RELU activation function in this layer. Then, we apply Gaussian diffusion to get diffused $\overline{\mu}$ and $\overline{log\sigma}$. Furthermore, we sample the posterior $\mathbf{Z}$ by reparameterization, where $\epsilon \sim \mathcal{N}(0, I)$. After sampling, we add up two terms of element-wise multiplication, $\left( \mathbb{J} - \mathbf{W}_{z} \right) \mathbf{H}^{(1)}$ and $\mathbf{W}_{z} \mathbf{Z}$. During the training, we apply node embedding propagation to further strengthen robustness. In the end, we output the $\mathbf{H}^{(1)}$ as the node embedding $\mathbf{H}_{emb}$ and obtain the output node representation $\mathbf{H}_{out}$ by aggregating nodes to the second-layer node hidden representation $\mathbf{H}^{(2)}$. Note that the solid arrows denote the directions of the forward propagation, whereas the dashed arrows denote the node embedding propagation.}
\label{fig:fig_model}
\end{figure}

%\subsection{The Proposed Model}
%\label{subsec:model}
\noindent
\textbf{\large The Proposed Model.}
Variational sampling is widely used in variational graph auto-encoders to generate node representation for graph reconstructions \cite{kipf2016variational, pan2018adversarially}. We apply Gaussian diffusion to variational sampling along with node embedding propagation to mitigate the above issues in our model. The overall framework is briefly introduced in Fig.~\ref{fig:fig_model}. Specifically, given a graph $\mathcal{G}$=$(\mathbf{\hat{A}}, \mathbf{X})$, our model first generates the node hidden representation $\mathbf{H}^{(1)}$, the matrix of mean $\mu$, and the matrix of log standard deviation $log\sigma$ by Eq.~\ref{eqn:mpm} as follows:
\begin{equation}
\left\{
  \begin{array}{lr}
    \mathbf{H}^{(1)} = ReLU \left( \textit{f}_{\mathbf{W}_{h}^{(0)}} \left( \mathbf{\hat{A}}, \mathbf{X} \right) \right), & \\
    \mu = ReLU \left( \textit{f}_{\mathbf{W}_{\mu}} \left( \mathbf{\hat{A}}, \mathbf{X} \right) \right), & \\
    log\sigma = ReLU \left( \textit{f}_{\mathbf{W}_{\sigma}} \left( \mathbf{\hat{A}}, \mathbf{X} \right) \right),
  \end{array}
\right.
\label{eqn:layer1}
\end{equation}
where $\mathbf{W}_{h}^{(0)} \in \mathbb{R}^{d \times h}$, $\mathbf{W}_{\mu} \in \mathbb{R}^{d \times h}$, and $\mathbf{W}_{\sigma} \in \mathbb{R}^{d \times h}$ represent the corresponding weight matrix for $\mathbf{H}^{(1)}$, $\mu$, and $log\sigma$. $ReLU(\cdot)$ is a non-linear activation function.

Inspired by the variational diffusion models~\cite{ho2020denoising, kingma2021variational}, we apply Gaussian diffusion to the $\mu$ and $log\sigma$ matrices with the decreasing diffusion rate $\gamma$ as follows:
\begin{equation}
\left\{
  \begin{array}{lr}
    \overline{\mu} = \sqrt{\Gamma_{t}}\mu, & \\
    \overline{log\sigma} = \sqrt{( 1 - \Gamma_{t})} log\sigma,
  \end{array}
\right.
\label{eqn:df}
\end{equation}
where $\overline{\mu}$ and $\overline{log\sigma}$ are denoted as the diffused $\mu$ and $log\sigma$, respectively; $\Gamma_{t} = \prod_{i=1}^{t} \gamma_{i}$ is the accumulated production of previous diffusion rates in the $t$-th iteration. The $\gamma$ linearly decreases with each iteration. Such a decrease simulates the process that a drop of ink spreads on the water.

To ensure that the gradient can be back-propagated, we sample the node hidden representation $\mathbf{Z}$ from $\overline{\mu}$ and $\overline{log\sigma}$ using a reparameterization trick.
\begin{equation}
\mathbf{Z} = \overline{\mu} + \epsilon e^{\overline{log\sigma}}, \ \  \epsilon \sim \mathcal{N}(0, I).
\label{eqn:pt}
\end{equation}

After sampling, we add up two terms of element-wise multiplication, $\left( \mathbb{J} - \mathbf{W}_{z} \right) \mathbf{H}^{(1)}$ and $\mathbf{W}_{z} \mathbf{Z}$, where $\mathbb{J} \in \mathbb{R}^{|\mathbf{Z}| \times h}$ denotes the unit matrix (a.k.a. all-ones matrix) and the weight matrix $\mathbf{W}_{z} \in \mathbb{R}^{|\mathbf{Z}| \times h}$ can be automatically learned through back-propagation. Our experiments empirically verify that such an approach can mitigate the deterioration of GNNs that we presented in the ablation study.
\begin{equation}
\mathbf{H}^{(1)} = \left( \mathbb{J} - \mathbf{W}_{z} \right) \mathbf{H}^{(1)} + \mathbf{W}_{z} \mathbf{Z}.
\label{eqn:h_z}
\end{equation}

To further strengthen robustness, we apply node embedding propagation during the training on inaccurately predicted nodes, whose neighbors are sampled in the preceding iteration. Algo.~\ref{algo:topo_sampling} presents the procedure by which we find the neighbors of the nodes with inaccurately predicted labels $\mathbf{\hat{Y}}^{t}$, s.t., $\mathbf{\hat{Y}}^{t} \neq \mathbf{Y}^{t}$, in the $t$-th iteration. To ascertain the neighbors of each inaccurately predicted node, we first sample a node label $\mathbf{\bar{y}}^{t}_{i}$ of the $i$-th node $v_i$ using a topology-based label sampler ({\bf line 3}) and then select the neighbors with the same label of $\mathbf{\bar{y}}^{t}_{i}$ ({\bf line 4}).

\begin{algorithm}
\caption{{\sc Topology}-based sampling for selecting neighbors of inaccurate nodes in the $t$-th iteration.}
\label{algo:topo_sampling}
\textbf{Input}: Predicted labels $\mathbf{\hat{Y}}^{t}$, The number of train nodes $N_{tr}$, Topology-based label sampler.
\begin{algorithmic}[1] %[1] enables line numbers
\FOR{$i \gets 0$  $\textbf{to}$  $N_{tr}$}
    \IF{$\mathbf{\hat{y}}^{t}_{i}$ is inaccurate}
        \STATE{Sample $\mathbf{\bar{y}}^{t}_{i}$ with the topology-based label sampler;}
        \STATE{Select the neighbors of $v_{i}$ based on $\mathbf{\bar{y}}^{t}_{i}$.}
    \ENDIF
\ENDFOR
\end{algorithmic}
\end{algorithm}

In the next iteration, for each inaccurately predicted node, we compute the mean matrix of the node embedding from its selected neighbors and then replace the node embedding of this node with the mean matrix. The benefit is that training the model with the propagated node embedding can effectively mitigate the vulnerability of these inaccurate nodes.

In the end, our model outputs $\mathbf{H}^{(1)}$ as the node embedding $\mathbf{H}_{emb}$ and $\mathbf{H}^{(2)}$ as the output node representation $\mathbf{H}_{out}$ for the downstream node classification. We compute $\mathbf{H}^{(2)}$ by:
\begin{equation}
\mathbf{H}^{(2)} = \textit{f}_{\mathbf{W}_{h}^{(1)}} \left( \mathbf{\hat{A}}, \mathbf{H}^{(1)} \right),
\label{eqn:layer2}
\end{equation}
where $\mathbf{W}_{h}^{(1)} \in \mathbb{R}^{h \times K}$ denotes the weight matrix for $\mathbf{H}_{out}$.

\noindent
{\bf \large Optimization.} 
%To learn robust node embedding, 
We optimize our model with the following loss functions.
First, we compute the cross entropy between predicted labels and ground-truth labels in the train graph.
\begin{equation}
%\scriptsize
\mathcal{L}_{ce} = -\frac{1}{N_{tr}} \sum_{i=1}^{N_{tr}} \sum_{k=1}^{K} y_{i,k}log(\hat{y}_{i,k}),
\label{eqn:loss_ce}
\end{equation}
where $y_{i,k}$ denotes a ground-truth label of the $i$-th node in the $k$-th class; $\hat{y}_{i,k}$ denotes a predicted label of the $i$-th node in the $k$-th class; $N_{tr}$ denotes the number of train nodes; $K$ denotes the number of classes. In this work, we compute the $i$-th predicted label $\hat{y}_{i}$ by selecting the maximum probability of the corresponding categorical distribution, which is computed by employing a softmax function on the $i$-th $\mathbf{H}_{out}$.

To obtain the $\mu$ and $log\sigma$ for subsequent sampling, we apply the Kullback-Leibler divergence between the posterior distribution $\mathcal{Q}(\mu, \sigma^2)$ before diffusion, where $\sigma =e^{log\sigma}$, and the prior distribution $\mathcal{N}(0, I)$ to ensure that these two distributions are as close as possible.
\begin{equation}
\mathcal{L}_{kl} = KL\left[ \mathcal{Q}(\mu, \sigma^2) \| \mathcal{N}(0, I) \right].
\label{eqn:loss_kl}
\end{equation}

After applying diffusion by Eq.~\ref{eqn:df}, we further match the sampled distribution $\mathbf{Z} \sim \mathcal{Q}(\overline{\mu}, \overline{\sigma}^2)$ with the Gaussian noise $\epsilon \sim \mathcal{N}(0, I)$ for preserving the Markovian property, where $\overline{\sigma} = e^{\overline{log\sigma}}$.
\begin{equation}
\mathcal{L}_{df} = \mathbb{E} \left[ \| \epsilon - \mathbf{Z} \|^2 \right].
\label{eqn:loss_df}
\end{equation}

During the retraining process, we match the predicted node embedding $\mathbf{\hat{H}}_{emb}$ (on the perturbed graphs) with the given pre-trained node embedding $\mathbf{H}_{emb}$ (on the unperturbed graphs) to enforce $\mathbf{\hat{H}}_{emb}$ as identically as $\mathbf{H}_{emb}$.
\begin{equation}
\mathcal{L}_{nm} = \mathbb{E} \left[ \| \mathbf{H}_{emb} - \mathbf{\hat{H}}_{emb} \|^2 \right].
\label{eqn:loss_nm}
\end{equation}

Overall, we formulate the total objective function as:
\begin{equation}
\mathcal{L}_{GVDN} = \lambda^{T} \mathcal{L},
\label{eqn:loss_total}
\end{equation}
where $\mathcal{L} = [\mathcal{L}_{ce}, \mathcal{L}_{kl}, \mathcal{L}_{df}, \mathcal{L}_{nm}]$ is a vector of losses; $\lambda^{T} \in \mathbb{R}^{|\mathcal{L}| \times 1}$ is a transposed vector of the balanced parameters, which is used for balancing the scale of each loss. Note that we only utilize the first three loss functions, $\mathcal{L}_{ce}$, $\mathcal{L}_{kl}$, and $\mathcal{L}_{df}$, during the training process and further include the last one, $\mathcal{L}_{nm}$, for retraining purposes.

\noindent
{\bf \large Training.}
The procedure is summarized in Algo.~\ref{algo:training}. As an input, we use the ground-truth labels $\mathbf{Y}$ as the train labels during training, whereas, for retraining, we use the pseudo labels $\mathbf{\hat{Y}}_{pseudo}$.
In each iteration, we compute losses, $\mathcal{L}_{ce}$, $\mathcal{L}_{kl}$, and $\mathcal{L}_{df}$ after the forward propagation during the training ({\bf line 2-5}) and additionally compute the node-embedding matching loss $\mathcal{L}_{nm}$ for retraining purposes ({\bf line 6-7}). Note that node embedding $\mathbf{H}_{emb}$ is used in both Eq.~\ref{eqn:loss_nm} and Algo.~\ref{algo:retrain} during retraining. Besides, we select the neighbors of inaccurately predicted nodes with the topology-based sampler using Algo.~\ref{algo:topo_sampling} for node embedding propagation in the next iteration ({\bf line 9}). At the end of each iteration, we apply backward propagation to compute the gradient ({\bf line 10}) and then update all necessary weights ({\bf line 11}). This procedure repeats $N_{epo}$ times until the loss has converged.

\begin{algorithm}[h]
\caption{{\sc Training} procedure of our model.}
\label{algo:training}
\textbf{Input}: Graph $\mathcal{G}_{train} = (\mathbf{\hat{A}}_{train}, \mathbf{X}_{train})$, train labels, Node aggregator, Topology-based sampler, The number of training epochs $N_{epo}$.
\begin{algorithmic}[1]
\FOR{$t \gets 1$  $\textbf{to}$  $N_{epo}$}
  \STATE{$\mathbf{H}_{out} \gets$ Forward propagation;}
  \STATE{$\mathcal{L}_{ce} \gets$ Compute the cross entropy loss by Eq.~\ref{eqn:loss_ce};}
  \STATE{$\mathcal{L}_{kl} \gets$ Compute the KL-divergence by Eq.~\ref{eqn:loss_kl};}
  \STATE{$\mathcal{L}_{df} \gets$ Compute the diffusion loss by Eq.~\ref{eqn:loss_df};}
  \IF{Retraining}
    \STATE{$\mathcal{L}_{nm} \gets$ Compute the node-embedding matching loss by Eq.~\ref{eqn:loss_nm};}
  \ENDIF
  \STATE{Select the neighbors of inaccurately predicted nodes with the topology-based sampler by Algo.~\ref{algo:topo_sampling};}
  \STATE{Backward propagation using Eq.~\ref{eqn:loss_total};}
  \STATE{Update the weights $\mathbf{W}_{h}^{(0)}$, $\mathbf{W}_{\mu}$, $\mathbf{W}_{\sigma}$, $\mathbf{W}_{z}$, and $\mathbf{W}_{h}^{(1)}$.}
\ENDFOR
\end{algorithmic}
\end{algorithm}

\begin{algorithm}[h]
\caption{{\sc Retraining} mechanism.}
\label{algo:retrain}
\textbf{Input}: A trained model, Node embedding $\mathbf{H}_{emb}$, Graph $\mathcal{G}_{pert} = (\mathbf{\hat{A}}_{pert}, \mathbf{X}_{pert})$.
\begin{algorithmic}[1] %[1] enables line numbers
\STATE{$\mathbf{\hat{Y}}_{pseudo} = \mathop{\arg \max} \text{Softmax} \left( \textit{f}_{\mathbf{W}_{h}^{(1)}} \left( \mathbf{\hat{A}}_{pert}, \mathbf{H}_{emb} \right) \right)$;}
\STATE{Retrain the model with $\mathbf{\hat{Y}}_{pseudo}$ by Algo.~\ref{algo:training}.}
\end{algorithmic}
\end{algorithm}

\noindent
{\bf \large Retraining Mechanism.}
In this work, we design a retraining mechanism for recovering the performance of the GNN on perturbed graphs. As presented in Algo.~\ref{algo:retrain}, we first generate pseudo labels $\mathbf{\hat{Y}}_{pseudo}$ using the given node embedding $\mathbf{H}_{emb}$ on perturbed graphs ({\bf line 1}) and then retrain the model with $\mathbf{\hat{Y}}_{pseudo}$ by Algo.~\ref{algo:training} ({\bf line 2}). Overall, we train the model to generate robust node embedding $\mathbf{H}_{emb}$, which can be used for recovering the performance of GNNs on perturbed graphs via our retraining mechanism.

\noindent
{\bf \large Time Complexity.}
During the training, we train our model with $N_{epo}$ epochs. For each epoch, our model traverses the number of train nodes $N_{tr}$ for forward and backward propagation. Besides, our model traverses the number of inaccurately predicted nodes $N_{ip}$ in the training nodes to sample the corresponding neighbors for node embedding propagation in the next iteration, where $N_{ip} \leq N_{tr}$. Thus, the {\bf node-level time complexity} is $\mathcal{O}(N_{epo} \cdot N_{tr})$. To consider the time complexity at the element level, we can further multiply by $f_l$, where $f_l$ is the dimensionality of the $l^{th}$ hidden layer. So, the {\bf element-level time complexity} is $\mathcal{O}(N_{epo} \cdot N_{tr} \cdot f_l)$.

%\section{Experiments}
\section{Experiments}
\label{sec:exp}
In this section, we conduct ablation studies, generalization tests, comparison of competing methods, visualizations of node embeddings, analysis of convergence and parameters, and further discuss limitations and future directions.

\noindent
{\bf \large Experimental Settings.}
We validate our model over six public graph datasets and follow the setting from~\cite{zhuang2022robust} to split the train, validation, and test graph to 10\%, 20\%, and 70\% of the nodes, respectively. We compare our model with nine competing methods and evaluate the performance in accuracy (Acc.) and average normalized entropy (Ent.).
Details are described in the {\bf Appendix}.

\begin{figure}[t]  % EX1-1
  %\hfill
  \begin{subfigure}{0.5\linewidth}
    \centering % include the first image
    \includegraphics[width=\textwidth]{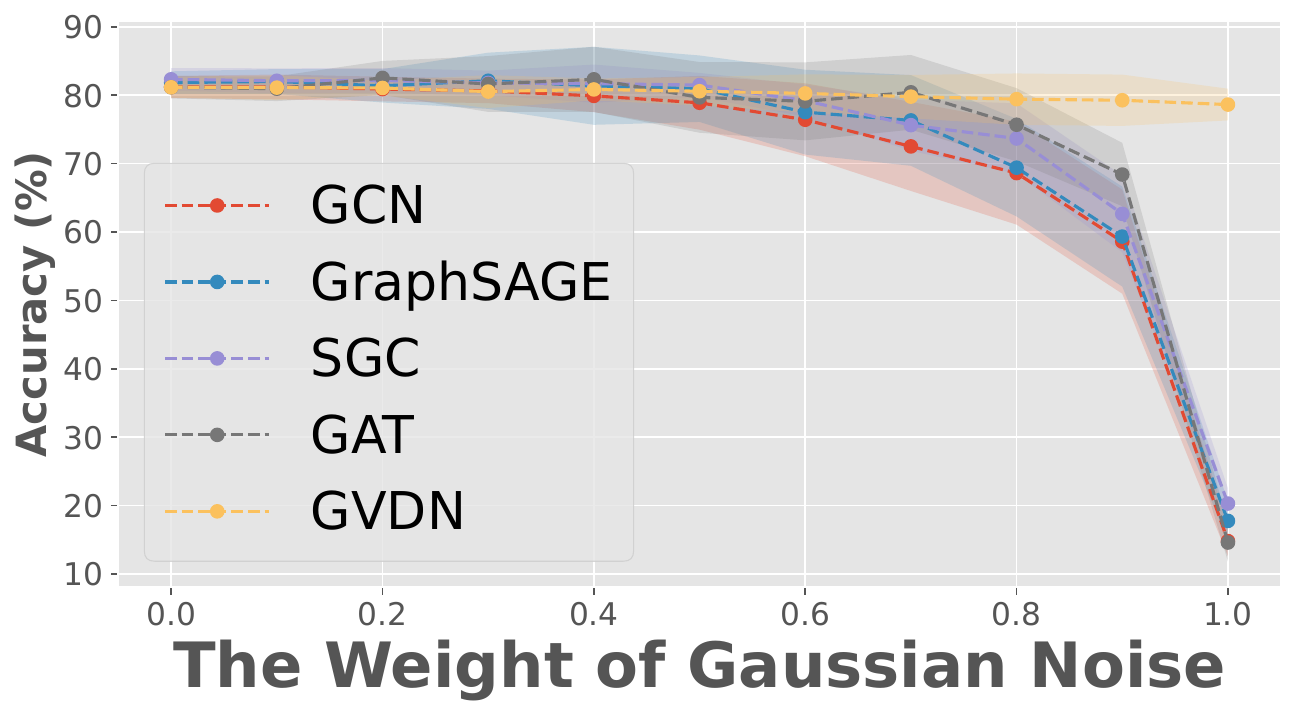}
  \end{subfigure}%
  %\hfill
  \begin{subfigure}{0.5\linewidth}
    \centering % include the second image
    \includegraphics[width=\textwidth]{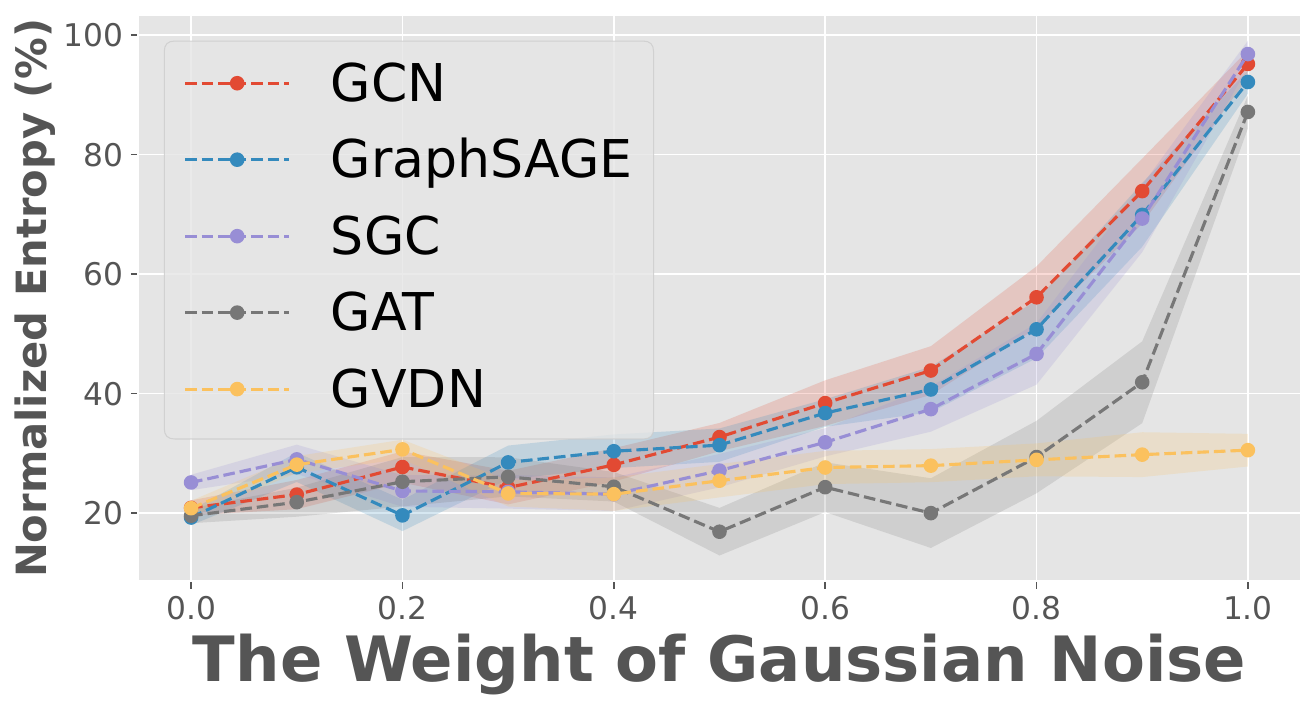}
  \end{subfigure}
\caption{Verification of effectiveness on our vanilla model architecture by investigating how Gaussian noise affects the performance of the node classifications.}
\label{fig:fig_wz}
\end{figure}

\noindent
{\bf \large Ablation Study.}
We conducted three experiments to examine our model architecture.
First, we investigate how GNNs behave as the weight of the Gaussian noise $\mathbf{W}_{ns}$ (scalar) in the hidden layers increases and also compare the performance on Cora over four representative GNN models, including GCN~\cite{kipf2016semi}, GraphSAGE~\cite{hamilton2017inductive}, SGC~\cite{wu2019simplifying}, and GAT~\cite{velivckovic2018graph}, with our model. 
Specifically, we add standard Gaussian noise $\mathcal{N}(0, I)$ in the $l$-th hidden layer of four GNNs with the weight as follows:
\begin{equation}
\mathbf{H}^{(l)} = \left( 1 - \mathbf{W}_{ns} \right) \mathbf{H}^{(l)} + \mathbf{W}_{ns} \mathcal{N}(0, I).
\label{eqn:gn}
\end{equation}
We adjust the weights of four GNNs and present the trends in Fig.~\ref{fig:fig_wz}. The trends show that the accuracy decreases as the percentage of the weight increases and even drops dramatically when the percentage is higher than 90\%. The normalized entropy has a similar pattern, which increases as the percentage of the weight increases.

\noindent
$\bullet$ {\bf Observation}: Simply increasing the percentage of the Gaussian noise in the hidden layer of GNNs by Eq.~\ref{eqn:gn} will deteriorate the performance of node classifications, i.e., lead to lower accuracy and higher normalized entropy.

\noindent
$\bullet$ {\bf Understanding}: Node aggregation by Eq.~\ref{eqn:mpm} can be regarded as a kind of low-pass filtering (a.k.a. Gaussian filtering) because the message-passing mechanism aggregates the node features from local neighbors~\cite{nt2019revisiting}. Increasing the percentage of Gaussian noise in the hidden space tends to over-smooth the hidden representation and thus make node embeddings less distinct, which is well-known as an over-smoothing issue, i.e., distinct nodes have similar node embeddings in hidden spaces~\cite{bo2021beyond}. This observation motivates us to explore a sophisticated approach that not only addresses over-smoothing issues but also leverages Gaussian noise against perturbation.

On the other hand, in the vanilla architecture of our model, we only concatenate both $\mathbf{H}^{(1)}$ and $\mathbf{Z}$ with the weight $\mathbf{W}_z$ by Eq.~\ref{eqn:h_z} without applying Gaussian diffusion and node embedding propagation. In our model, adjusting the weight of $\mathbf{Z}$ could be equivalent to changing the percentage of Gaussian noise in the hidden layers. The results in Fig.~\ref{fig:fig_wz} show that our vanilla architecture can effectively maintain the performance, i.e., higher Acc. and lower Ent., as the percentage of the weight increases. This is possible since our model can learn the weight and adjust the values of noise to different nodes instead of assigning the same value to all nodes.

\begin{figure}[t]  % EX1-2
  %\hfill
  \raisebox{0.4\height}{\rotatebox{90}{\scriptsize $p_{rdm}=1\%$}}
  \begin{subfigure}{0.46\linewidth}
    \centering % include the first image
    \includegraphics[width=\textwidth]{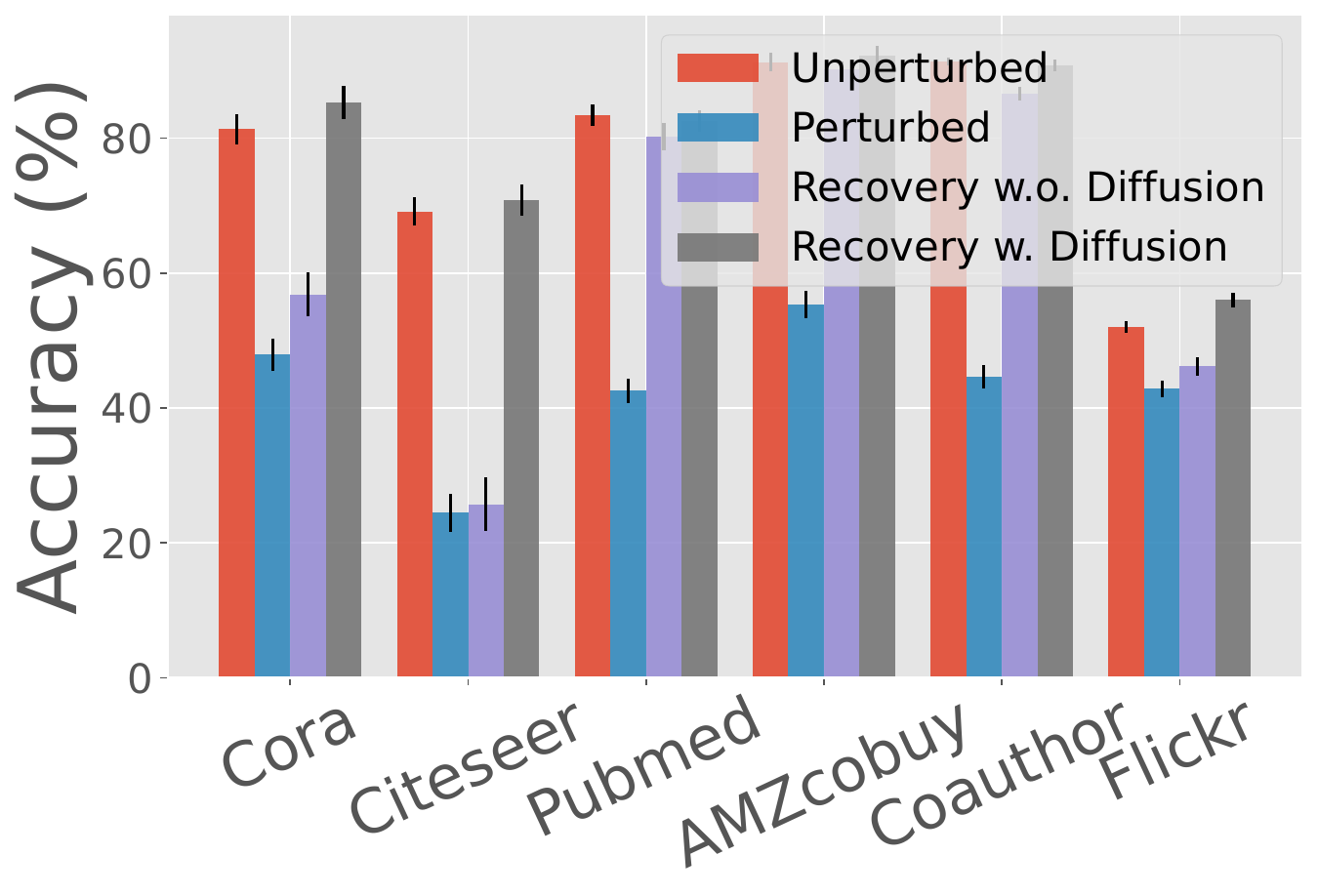}
    %\caption{$p_{rdm}=1\%$}
  \end{subfigure}%
  %\hfill
  \raisebox{0.4\height}{\rotatebox{90}{\scriptsize $p_{rdm}=5\%$}}
  \begin{subfigure}{0.46\linewidth}
    \centering % include the second image
    \includegraphics[width=\textwidth]{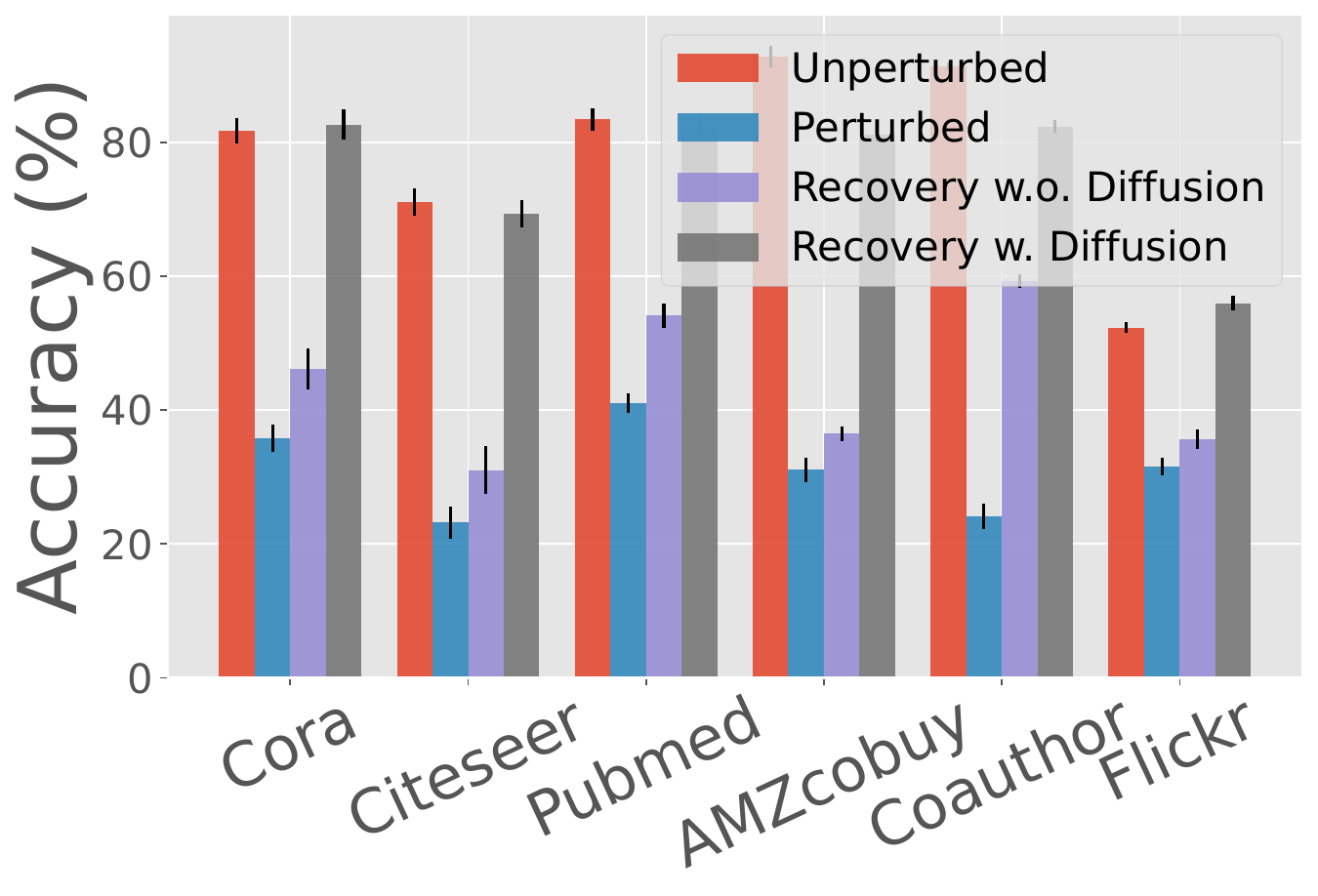}
    %\caption{$p_{rdm}=5\%$}
  \end{subfigure}
\caption{Ablation study of Gaussian diffusion. $p_{rdm}$ denotes the percentage of random-perturbator nodes over victim nodes. ``Unperturbed" and ``Perturbed" denote the accuracy on corresponding graphs, whereas ``Recovery" denotes the recovered accuracy on perturbed graphs. Results affirm that applying Gaussian diffusion can generalize robust recovery to all perturbed graphs.}
\label{fig:fig_diffusion}
\end{figure}

\begin{figure}[t] % EX1-3
  \centering
  \includegraphics[width=\linewidth]{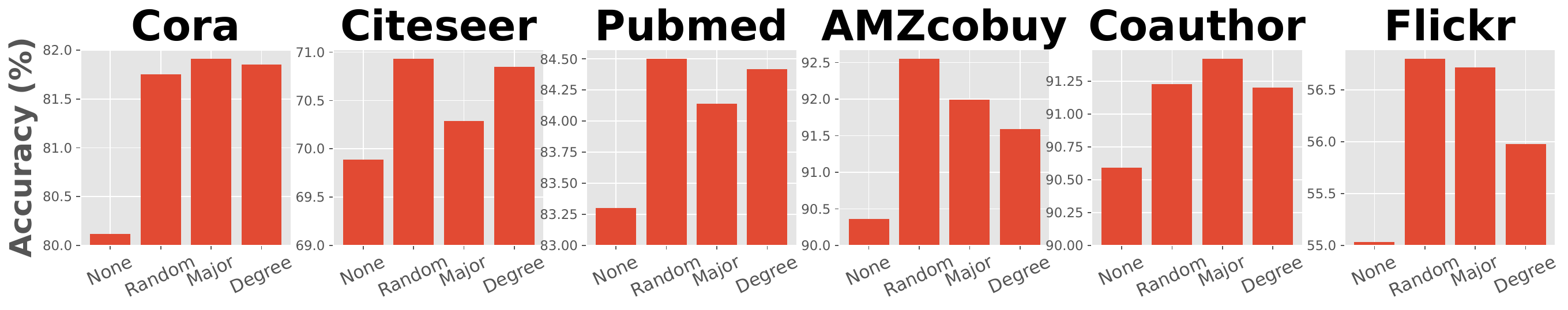}
  \caption{Investigation on the node embedding propagation with three types of samplers (random, major, and degree). ``None" denotes the performance without using node embedding propagation.}
\label{fig:fig_kernels}
\end{figure}

Based on the vanilla architecture, we further examine whether Gaussian diffusion can generalize robust recovery on six perturbed graphs (random perturbations) under two different percentages of random-perturbator nodes over victim nodes (1\% and 5\%).
As shown in each sub-figure of Fig.~\ref{fig:fig_diffusion}, the first two bars display the accuracy on unperturbed and perturbed graphs, whereas the last two bars present the recovered accuracy without/with applying Gaussian diffusion.
This ablation study validates the effectiveness of Gaussian diffusion. Even in a gentle case, e.g., $p_{rdm}$=1\%, our vanilla model cannot generalize robust recovery on Cora and Citeseer without applying Gaussian diffusion. In brief, the results indicate that applying Gaussian diffusion to the vanilla architecture can enhance the generalization capability of robust recoveries.
Moreover, we dive deeper into the node embedding propagation with three types of samplers \cite{zhuang2022robust} based on our previous architecture, applying Gaussian diffusion to the vanilla architecture. According to Fig.~\ref{fig:fig_kernels}, propagating node embedding during training can further strengthen robustness. All three samplers are effective in node embedding propagation.

\begin{table*}[t] % EX2
\scriptsize
%\footnotesize
%\small
\centering
\setlength{\tabcolsep}{2pt}
\begin{tabular}{c|cc|cc|cc|cc|cc|cc} % rows=1+2*3, cols=1+2*5
\toprule % Title
\multirow{2}{*}{\parbox{1.3cm}{\centering \textbf{Methods}}} & 
  \multicolumn{2}{c|}{\textbf{Cora}} &
  \multicolumn{2}{c|}{\textbf{Citeseer}} &
  \multicolumn{2}{c|}{\textbf{Pubmed}} &
  \multicolumn{2}{c|}{\textbf{AMZcobuy}} &
  \multicolumn{2}{c|}{\textbf{Coauthor}} &
  \multicolumn{2}{c}{\textbf{Flickr}}
  \\
\cline{2-13}
  & \textbf{Acc.} & \textbf{Ent.} & \textbf{Acc.} & \textbf{Ent.} & \textbf{Acc.} & \textbf{Ent.} & \textbf{Acc.} & \textbf{Ent.} & \textbf{Acc.} & \textbf{Ent.} & \textbf{Acc.} & \textbf{Ent.}  \\
\midrule
{\bf clean} & 81.43 (0.78) & 26.47 (2.35) & 71.06 (0.54) & 61.53 (3.64) & 84.34 (0.63) & 37.49 (2.56) & 92.75 (0.97) & 14.22 (2.07) & 91.50 (0.89) & 8.82 (1.11) & 52.06 (0.85) & 66.29 (3.50) \\
\midrule
{\bf rdmPert} & 48.42 (2.99) & {\bf 5.98 (1.14)} & 24.46 (2.66) & {\bf 6.06 (1.34)} & 42.80 (1.94) & {\bf 1.81 (0.75)} & 55.29 (2.86) & 21.85 (1.17) & 44.24 (1.85) & 17.21 (2.99) & 42.86 (1.22) & 69.66 (3.91) \\
{\bf retrain} & 82.63 (2.35) & 29.04 (2.27) & {\bf 69.96 (2.50)} & 55.11 (3.01) & {\bf 82.11 (1.41)} & 35.44 (2.85) & {\bf 92.17 (1.19)} & 10.82 (1.11) & {\bf 89.80 (1.25)} & 10.28 (1.86) & {\bf 56.80 (1.09)} & 7.96 (1.20) \\
\midrule
{\bf infoSparse} & 72.26 (2.27) & 36.49 (2.94) & 64.10 (2.73) & 68.11 (3.50) & 76.58 (1.61) & 47.37 (3.33) & 90.34 (1.37) & 17.19 (1.47) & 86.91 (1.52) & 16.33 (2.71) & 46.33 (1.01) & 76.53 (3.70) \\
{\bf retrain} & 73.42 (2.04) & 29.77 (2.75) & 64.46 (2.81) & 32.74 (2.11) & 77.56 (1.53) & 35.99 (2.12) & 91.39 (1.45) & {\bf 9.36 (1.30)} & 87.58 (1.65) & 7.42 (1.33) & 49.71 (1.44) & {\bf 6.99 (1.18)} \\
\midrule
{\bf advAttack} & 29.44 (2.60) & 6.06 (1.59) & 4.05 (1.53) & 12.50 (1.19) & 1.77 (0.81) & 2.37 (0.94) & 76.05 (1.02) & 29.98 (2.16) & 61.03 (1.98) & 11.84 (1.82) & 45.20 (2.13) & 63.98 (2.80) \\
{\bf retrain} & {\bf 83.33 (2.54)} & 13.68 (2.49) & 59.64 (2.28) & 32.02 (2.67) & 79.20 (1.60) & 25.31 (3.35) & 77.17 (1.81) & 17.29 (1.64) & 63.38 (1.52) & {\bf 6.69 (0.97)} & 48.77 (2.09) & 10.66 (2.31) \\
\bottomrule
\end{tabular}
\caption{Generalization test on our proposed model under three types of perturbations across six datasets. ``clean" denotes the performance on the clean (unperturbed) graphs. ``rdmPert", ``infoSparse", and ``advAttack" denote the performances on the corresponding perturbed graphs. ``retrain" denotes the recovery in the corresponding scenario. We evaluate the performance five times ``mean (std)" by classification accuracy (Acc.) and average normalized entropy (Ent.) on victim nodes.}
\label{table:generalization}
\end{table*}

\begin{table*}[t] % EX3
\scriptsize
%\footnotesize
%\small
\centering
\setlength{\tabcolsep}{2.5pt}
\begin{tabular}{c|ccc|ccc|ccc|ccc|ccc|ccc} % rows=2+9, cols=1+3*5
\toprule % Title
\multirow{2}{*}{\parbox{1.3cm}{\centering {\bf Methods}}} & \multicolumn{3}{c|}{{\bf Cora}} & \multicolumn{3}{c|}{{\bf Citeseer}} & \multicolumn{3}{c|}{{\bf Pubmed}} & \multicolumn{3}{c|}{{\bf AMZcobuy}} & \multicolumn{3}{c|}{{\bf Coauthor}} & \multicolumn{3}{c}{{\bf Flickr}} \\
\cline{2-19}
& {\bf Acc.} & {\bf Ent.} & {\bf Time} & {\bf Acc.} & {\bf Ent.} & {\bf Time} & {\bf Acc.} & {\bf Ent.} & {\bf Time} & {\bf Acc.} & {\bf Ent.} & {\bf Time} & {\bf Acc.} & {\bf Ent.} & {\bf Time} & {\bf Acc.} & {\bf Ent.} & {\bf Time} \\
\midrule
{\bf GCN-Jaccard}~\cite{wu2019adversarial} & 69.47 & 88.91 & 2.41 & 57.94 & 89.83 & 2.28 & 67.13 & 77.29 & 5.63 & 89.15 & 91.32 & 8.87 & 86.99 & 96.23 & 9.38 & 49.37 & 95.19 & 53.15 \\
{\bf GCN-SVD}~\cite{entezari2020all} & 47.89 & 88.95 & 2.66 & 34.33 & 88.34 & 2.96 & 73.28 & 80.89 & 8.16 & 82.85 & 91.94 & 10.12 & 75.78 & 95.51 & 16.56 & 50.13 & 96.45 & 71.22 \\
{\bf DropEdge}~\cite{rong2019dropedge} & 78.42 & 90.40 & {\bf 1.36} & 46.61 & 93.35 & {\bf 2.51} & 67.29 & 62.22 & {\bf 3.20} & 52.52 & 94.55 & {\bf 3.18} & 54.83 & 97.66 & {\bf 4.80} & 41.50 & 99.09 & {\bf 11.69} \\
{\bf GRAND}~\cite{feng2020graph} & 51.26 & 89.69 & 4.76 & 31.45 & 92.29 & 3.74 & 49.32 & 89.47 & 11.06 & 40.09 & 95.38 & 11.30 & 58.02 & 96.69 & 129.82 & 41.58 & 95.99 & 233.41 \\
{\bf RGCN}~\cite{zhu2019robust} & 66.32 & 98.02 & 10.78 & 60.94 & 98.15 & 10.96 & 83.20 & 87.05 & 32.95 & 76.53 & 96.82 & 32.35 & 89.25 & 97.88 & 184.21 & 50.11 & 96.33 & 290.10 \\
{\bf ProGNN}~\cite{jin2020graph} & 47.37 & 87.17 & 202.01 & 34.89 & 77.19 & 340.68 & 49.25 & 85.93 & 3081.02 & 88.75 & 89.78 & 2427.02 & 82.16 & 95.85 & 3122.50 & 48.67 & 95.10 & 6201.90 \\
{\bf MC Dropout} (Gal 2016) & 77.37 & 32.46 & 5.16 & 66.52 & 70.27 & 7.81 & 79.80 & 59.44 & 14.37 & 89.91 & 19.65 & 8.66 & 89.49 & 16.91 & 34.79 & 50.60 &  60.53 & 11.81 \\
{\bf GraphSS} (Zhuang et al. 2022a) & 80.43 & {\bf 13.74} & 24.04 & 46.78 & {\bf 18.28} & 22.87 & 81.34 & {\bf 24.56} & 118.43 & 84.68 & 9.58 & 50.81 & 89.41 & {\bf 6.76} & 112.12 & 51.25 & 58.92 & 274.17 \\
{\bf LInDT} (Zhuang et al. 2022c) & 83.68 & 20.40 & 33.32 & 70.53 & 57.26 & 66.61 & 83.64 & 32.55 & 145.76 & 91.89 & {\bf 8.58} & 76.05 & 90.58 & 7.44 & 155.46 & 52.01 & 57.55 & 319.42 \\
{\bf Ours} & {\bf 84.01} & 29.57 & 21.30 & {\bf 71.49} & 53.75 & 21.60 & {\bf 84.25} & 34.16 & 105.44 & {\bf 92.56} & 10.06 & 40.96 & {\bf 91.22} & 9.85 & 95.01 & {\bf 57.15} & {\bf 8.91} & 211.01 \\
\bottomrule
\end{tabular}
\caption{Comparison between competing methods and our model under the random perturbations scenario. Acc. (\%) denotes classification accuracy. Ent. (\%) denotes average normalized entropy. Time (s) denotes total runtime.}
\label{table:competing}
\end{table*}

%\subsection{Generalization Test.}
\noindent
{\bf \large Generalization Test.}
We examine the generalization of our model (using the GCN aggregator with random sampler) under three types of perturbations, random perturbations (rdmPert), information sparsity (infoSparse), and adversarial attacks (advAttack), over six graphs, and present results in Tab.~\ref{table:generalization}.
We group the performance before/after retraining by the types of perturbations and highlight the highest Acc. and lowest Ent. on the results among these three groups.
The results indicate that our model can help recover the accuracy in all datasets. However, the recovery may get a lower margin on extremely sparse graphs. This is possible because node aggregators and node embedding propagation may fail to capture the information when the graph structure and the features are extremely sparsified. Note that we sparsify the graph before applying adversarial attacks on AMZcobuy, Coauthor, and Flickr.
For uncertainty, our model can reduce the normalized entropy (Ent.) in most cases compared to the Ent. on the clean graphs. In general, lower Ent. indicates lower uncertainty on the categorical distribution. However, this property is not always true when the graph is under severe perturbations. Worse prediction also leads to lower Ent., such as the results on Pubmed under advAttack (Acc.=1.77\%, Ent.=2.37\%). Overall, our goal is to achieve higher accuracy with relatively lower normalized entropy.

%\subsection{Comparison with Competing Methods}
\noindent
{\bf \large Comparison with Competing Methods.}
Tab.~\ref{table:competing} compares performance between competing methods and our model under rdmPert ($p_{rdm}$=1\%). The backbone model here is GCN. Both LInDT and our model use the random sampler. According to Tab.~\ref{table:competing}, our model outperforms all competing methods in Acc. and achieves low Ent. comparable to the state-of-the-art post-processing MCMC methods, LInDT. Compared among the post-processing methods (GraphSS, LInDT, and our model), MCMC methods (GraphSS and LInDT) can significantly decrease the Ent., whereas our model maintains a lower runtime than MCMC methods. Note that the main difference between MCMC methods and our model is that MCMC methods aim to approximate inferred labels to original labels as closely as possible on perturbed graphs, whereas our model is designed for generating node embedding for downstream tasks, which implies that our model has higher efficiency in the setting of streaming graphs (Further experiments are presented in {\bf Appendix}).

\begin{figure}[t]  % Visualization
  %\hfill
  \raisebox{0.6\height}{\rotatebox{90}{\bf Cora}}
  \begin{subfigure}{0.31\linewidth}
    \centering % include 1st image
    \includegraphics[width=\textwidth] {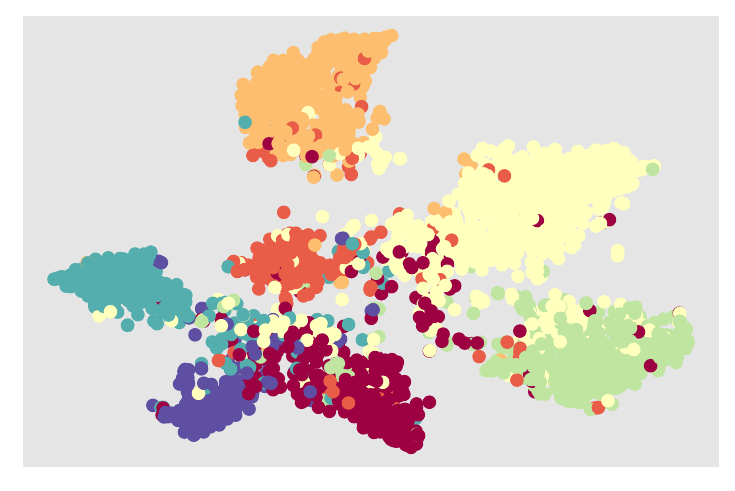}
  \end{subfigure}
  %\hfill
  \begin{subfigure}{0.31\linewidth}
    \centering % include 2nd image
    \includegraphics[width=\textwidth]{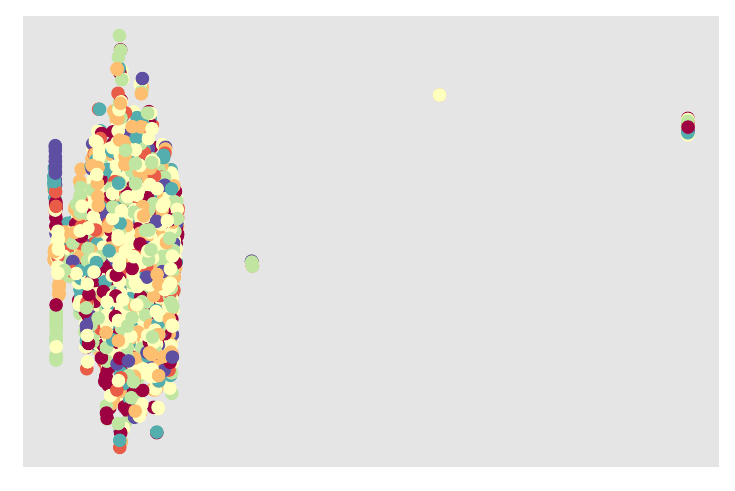}
  \end{subfigure}
  %\hfill
  \begin{subfigure}{0.31\linewidth}
    \centering % include 3rd image
    \includegraphics[width=\textwidth]{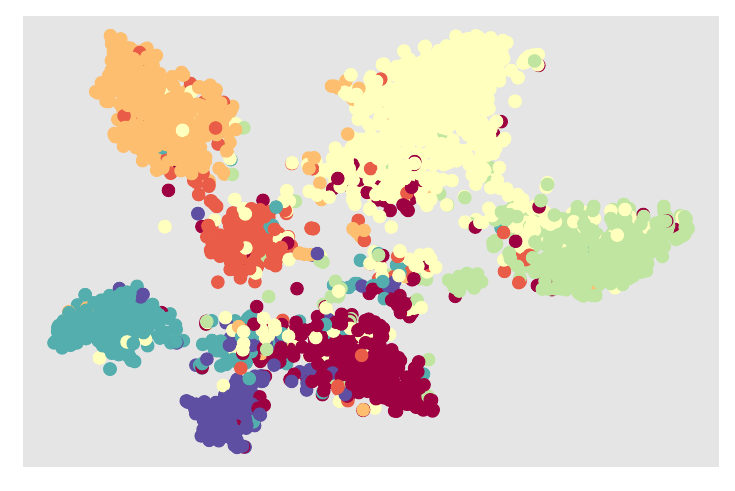}
  \end{subfigure}
  %\hfill
  \\
  \raisebox{0.2\height}{\rotatebox{90}{\bf Citeseer}}
  \begin{subfigure}{0.31\linewidth}
    \centering % include 4th image
    \includegraphics[width=\textwidth]{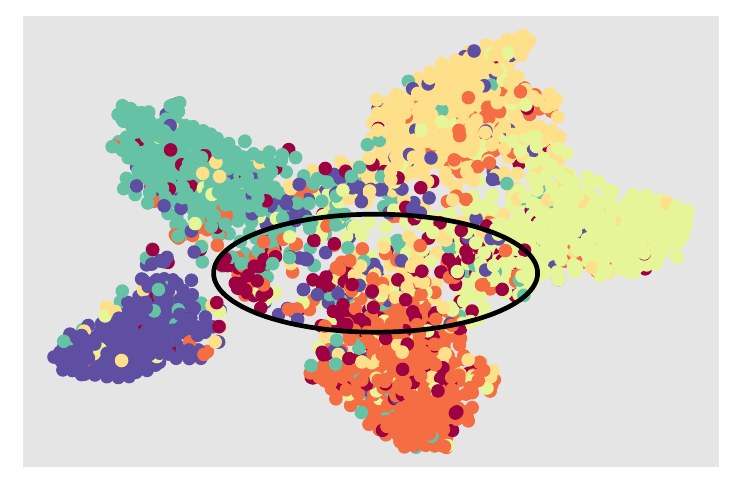}
  \end{subfigure}
  %\hfill
  \begin{subfigure}{0.31\linewidth}
    \centering % include 5th image
    \includegraphics[width=\textwidth]{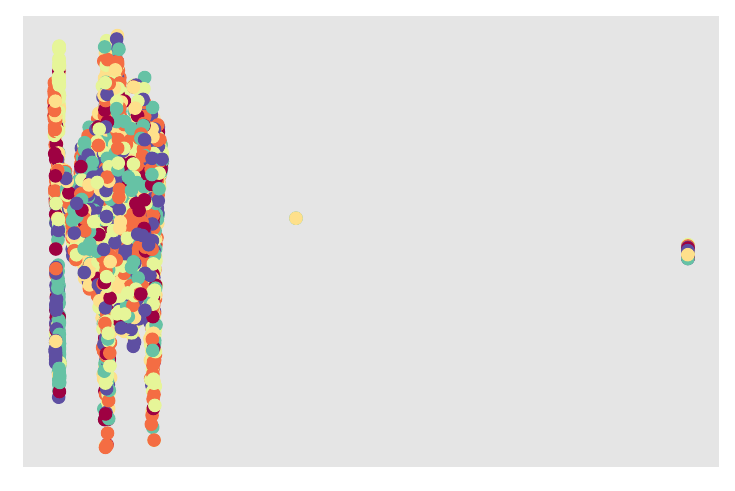}
  \end{subfigure}
  %\hfill
  \begin{subfigure}{0.31\linewidth}
    \centering % include 6th image
    \includegraphics[width=\textwidth]{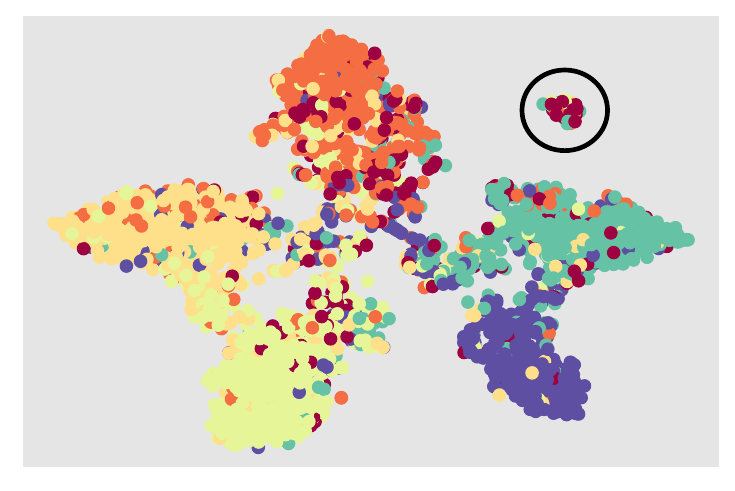}
  \end{subfigure}
\caption{Visualization of node embedding on an unperturbed graph ($1^{st}$ column), a perturbed graph under 1\% $p_{rdm}$ random perturbations ($2^{nd}$ column), and a perturbed graph after retraining ($3^{rd}$ column). Embeddings are colored by ground-truth labels.}
\label{fig:fig_tsne}
\end{figure}

\begin{figure}[t] % Recovery Process
  \centering
  \includegraphics[width=\linewidth]{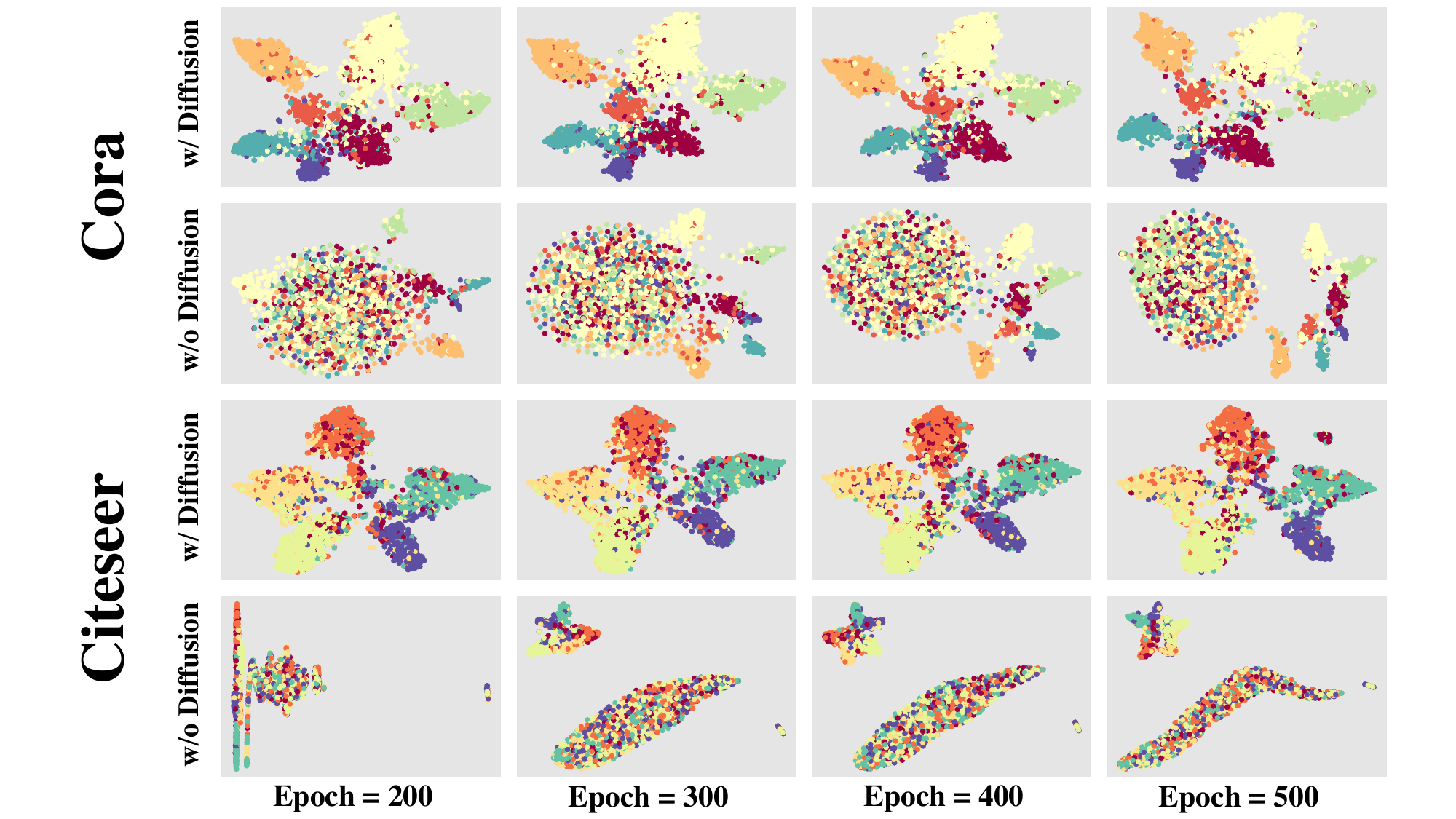}
  \caption{Visualization of the node embedding during the retraining process on two randomly perturbed graphs. ``w/ Diffusion" and ``w/o Diffusion" denote whether we apply Gaussian diffusion. ``Epoch" denotes the retraining epoch.}
\label{fig:fig_process}
\end{figure}

%\subsection{Visualization of Node Embedding}
\noindent
{\bf \large Visualization.}
%Besides examining the performance, 
We visualize the node embedding via t-SNE on Cora and Citeseer as examples in Fig.~\ref{fig:fig_tsne}. 
The visualization indicates that the retraining mechanism can help recover better node hidden representations in latent spaces on perturbed graphs. For example, in Citeseer (six classes), the nodes in the red class are mixed in the center ($1^{st}$ column) after initial training. Our proposed retraining mechanism can still partially classify these nodes in the upper-right corner ($3^{rd}$ column) under random perturbations.
To better understand the retraining process, we visualize node embeddings during retraining on two randomly perturbed graphs ($p_{rdm}$=1\%), Cora and Citeseer. The $1^{st}$ and $3^{rd}$ rows present results of applying Gaussian diffusion, whereas the $2^{nd}$ and $4^{th}$ rows present results without applying Gaussian diffusion. The retraining starts at the best checkpoint in the initial training process and stops by 500 epochs in total. We display a snapshot in the $200^{th}$, $300^{th}$, $400^{th}$, and $500^{th}$ epochs. The visualization validates the effectiveness of Gaussian diffusion.

\begin{figure}[t]  % Convergence
  %\hfill
  \begin{subfigure}{0.505\linewidth}
    \centering % include the 1st image
    \includegraphics[width=\textwidth]{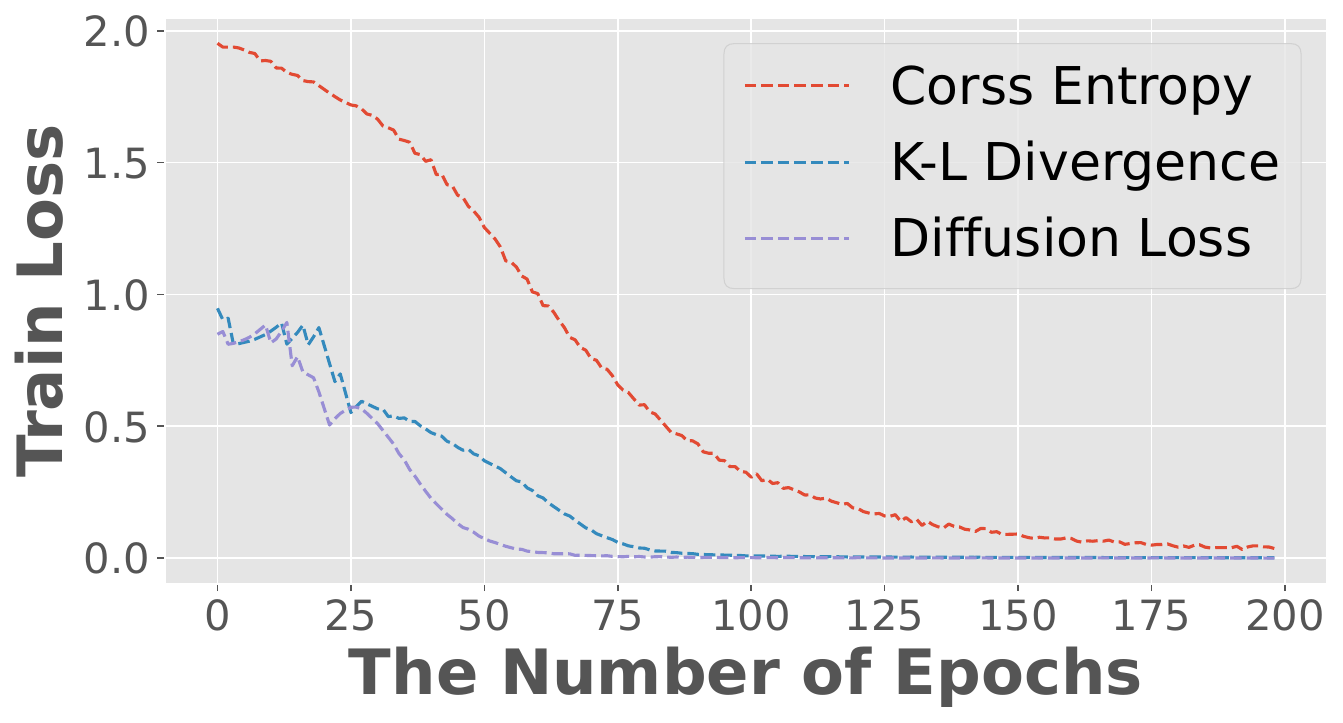}
  \end{subfigure}%
  %\hfill
  \begin{subfigure}{0.495\linewidth}
    \centering % include the 2nd image
    \includegraphics[width=\textwidth]{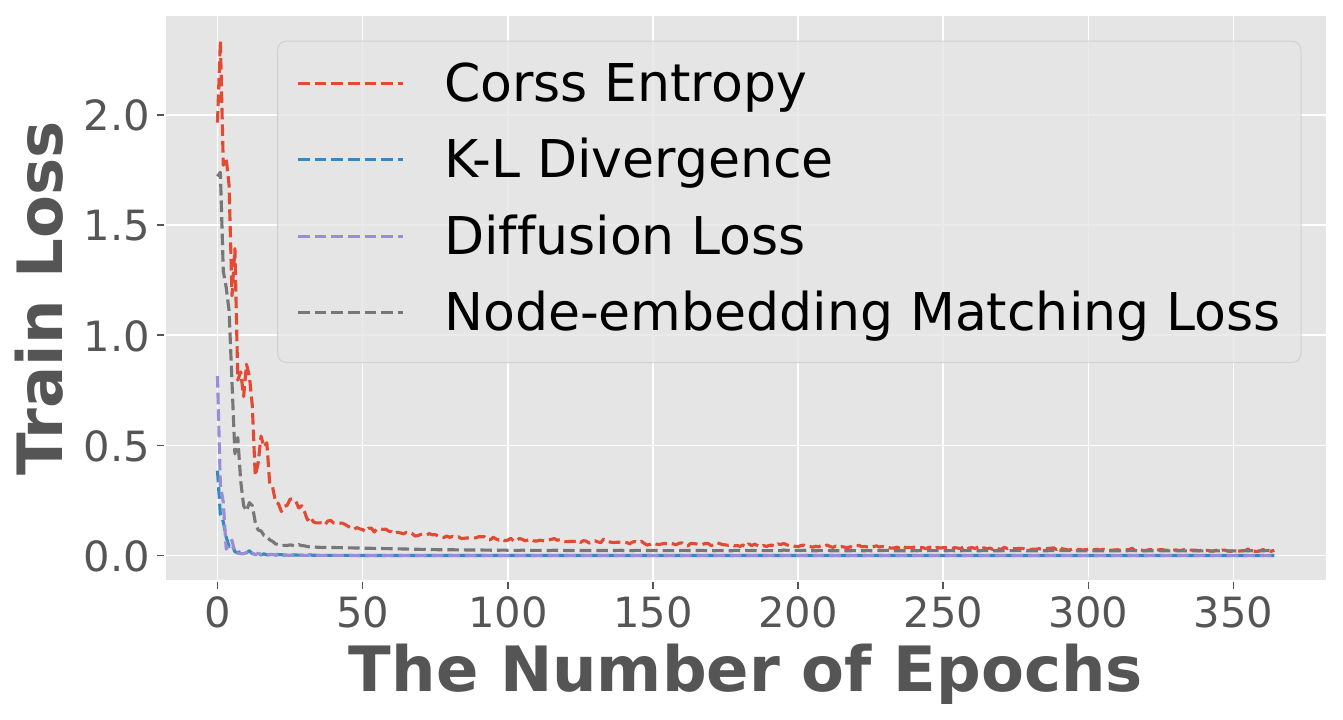}
  \end{subfigure}
\caption{Convergence verification for each loss function during training (left) and retraining (right).}
\label{fig:fig_loss}
\end{figure}

%\subsection{Verification of Convergence}
\noindent
{\bf \large Convergence.}
We present loss curves during training (left) and retraining (right) for verifying the convergence of each loss on Cora as an example in Fig.~\ref{fig:fig_loss}. For better presentation, we rescale the loss of K-L divergence and diffusion to the same magnitude of cross entropy. The results verify that all losses can eventually converge. Note that the retraining starts from the best checkpoint in the training stage and ends in 500 epochs (including the first 200 training epochs).

\begin{figure}[t] % dr
  \centering
  \includegraphics[width=\linewidth]{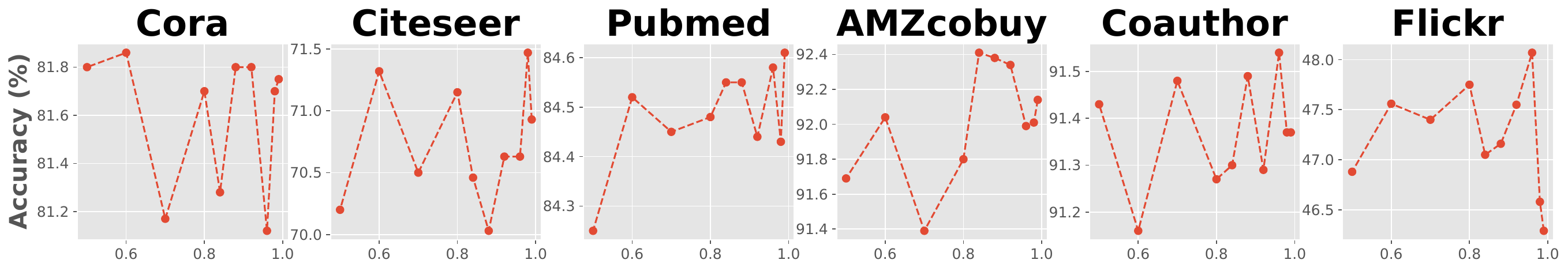}
  \caption{Analysis of diffusion rate (x-axis: minimum values of a range for diffusion rate; y-axis: validation accuracy).}
\label{fig:fig_params_dr}
\end{figure}

\begin{figure}[t] % lc
  \centering
  \includegraphics[width=\linewidth]{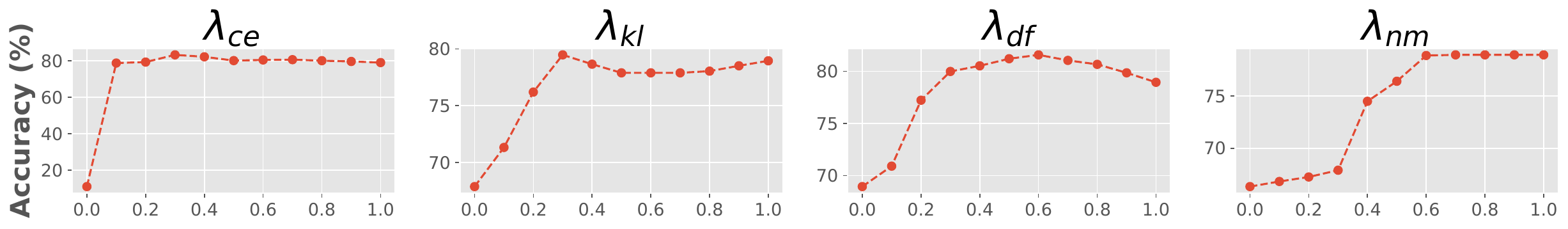}
  \caption{Investigation of coefficient sensitivity for each loss on Cora (x-axis: the value of $\lambda$; y-axis: validation accuracy).}
\label{fig:fig_params_lc}
\end{figure}

%\subsection{Analysis of Parameters}
\noindent
{\bf \large Parameters.}
%The diffusion rate $\gamma$ is a crucial parameter for Gaussian diffusion.
We tune the diffusion rate $\gamma$ within a given range on the validation set. The maximum value of the range $\gamma_{max}$ is 0.9999. The minimum value of the range $\gamma_{min}$ changes from 0.5 to 0.99. As displayed in Fig.~\ref{fig:fig_params_dr}, we select the $\gamma_{min}$ as 0.6, 0.98, 0.99, 0.84, 0.96, and 0.96 on six datasets.
Besides, we follow the same setting as that in Tab.~\ref{table:competing} and analyze the coefficient sensitivity for each loss by control variables on Cora as an example in Fig.~\ref{fig:fig_params_lc}. Cross-entropy is crucial to classification as accuracy drops dramatically after muting $\mathcal{L}_{ce}$. Increasing the percentage of $\mathcal{L}_{nm}$ can boost recovery. Choosing appropriate $\mathcal{L}_{kl}$ and $\mathcal{L}_{df}$ can help achieve higher accuracy. We assign equal weight to each loss and verify that our model can generalize to these public datasets without further tuning the coefficient.

%\subsection{Limitation and Future Directions}
\noindent
{\bf \large Limitation.}
Though we release dependency on a well-predicted prior compared to MCMC methods, our model still gets a lower margin on extremely sparse graphs due to a lack of information for node aggregation and propagation. In the future, we could utilize the inverse diffusion process to complete graph structures. Besides, node embedding propagation may not perform strongly on heterogeneous graphs due to the assumption of graph homophily property.

%\section{Related Work}
\section{Related Work}
\label{sec:rewk}

Besides discussing background in {\bf Introduction}, we further introduce related works as follows:
\noindent
{\bf Variational diffusion} models have achieved great impact as a kind of generative model in recent two years~\cite{song2020denoising, li2020variational, rey2019diffusion}. Following this impact, an increasing number of studies have started exploring how variational diffusion can contribute to graph structural data~\cite{zhou2020variational, ling2022source}. Learning robust node representation via variational diffusion is a promising direction in the early stage.
\noindent
{\bf Label propagation} techniques have been widely used in the semi-supervised learning paradigm~\cite{zhu2002learning, wang2007label, ugander2013balanced, gong2016label, iscen2019label}. Recent studies utilize this technique to learn robust GNNs~\cite{huang2020combining, wang2020unifying}.

\section{Conclusion}
\label{sec:con}
We aim to learn robust node representations for recovering node classifications on perturbed graphs. While infusing Gaussian noise in hidden layers is an appealing strategy for enhancing robustness, our experiments show that this strategy may result in over-smoothing issues during node aggregation. Thus, we propose the Graph Variational Diffusion Network (GVDN), a new node encoder that effectively manipulates Gaussian noise to improve robustness while alleviating over-smoothing issues by two mechanisms: Gaussian diffusion and node embedding propagation. We also design a retraining mechanism using generated node embeddings for recovery. Extensive experiments validate the effectiveness of our proposed model across six public graph datasets.

\section{Acknowledgments}
This work is partially supported by the NSF IIS-1909916.

\section{Broader Impact}
We confirm that we fulfill the author's responsibilities.
As an indispensable component of AI algorithms, GNNs are extensively utilized to learn node representations in graph-structure data. Thus, it is significant to improve the robustness of GNNs against perturbations. From a social perspective, our model can be applied to enhance the robustness in downstream node classification tasks in real-world scenarios, such as user profiling in online social networks. From a technical perspective, our model can benefit other research fields, such as robust node classifications in noise-label or sparse-label environments ({\bf Appendix}).

%\balance
\bibliography{2reference}
%\clearpage

\appendix
%\section{Appendix}
\section{APPENDIX}
\label{sec:appendix}

In the appendix, we first describe the details of datasets, implementation of perturbations, hyper-parameters for both our model and competing methods, evaluation metrics, and hardware/software. We further present supplementary experiments in the second section.

\section{Experimental Settings}

\subsection{Datasets}
%\noindent
%{\bf \large Datasets.}
Tab.~\ref{table:dataset} presents statistics of six public node-labeled graph datasets. \textbf{Cora}, \textbf{Citeseer}, and \textbf{Pubmed} are famous citation graph data~\citep{sen2008collective}. \textbf{AMZcobuy} comes from the photo segment of the Amazon co-purchase graph~\cite{shchur2018pitfalls}. \textbf{Coauthor} is co-authorship graphs of computer science based on the Microsoft Academic Graph from the KDD Cup 2016 challenge \footnote{https://www.kdd.org/kdd-cup/view/kdd-cup-2016}. \textbf{Flickr} is a image-relationship graph dataset~\cite{zeng2019graphsaint}. The first five graphs are widely used in the study of robust node representation learning. We additionally select Flickr to validate that our proposed model can achieve robust performance on a graph with a larger size and lower EHR.

\begin{table}[h] % Dataset
\footnotesize
%\scriptsize
\centering
\setlength{\tabcolsep}{3.8pt}
\begin{tabular}{ccccccc}
  \toprule
    \textbf{Dataset} & {$\left| \mathcal{V} \right|$} & {$\left| \mathcal{E} \right|$} & {$\left| F \right|$} & {$\left| C \right|$} & {Avg.D} & {EHR(\%)} \\
    \midrule
    {\bf Cora} & 2,708 & 10,556 & 1,433 & 7 & 4.99 & 81.00 \\
    {\bf Citeseer} & 3,327 & 9,228 & 3,703 & 6 & 3.72 & 73.55 \\
    {\bf Pubmed} & 19,717 & 88,651 & 500 & 3 & 5.50 & 80.24 \\
    {\bf AMZcobuy} & 7,650 & 287,326 & 745 & 8 & 32.77 & 82.72 \\
    {\bf Coauthor} & 18,333 & 327,576 & 6,805 & 15 & 10.01 & 80.81 \\
    {\bf Flickr} & 89,250 & 899,756 & 500 & 7 & 11.07 & 31.95 \\
  \bottomrule
\end{tabular}
\caption{Statistics of datasets. $\left| \mathcal{V} \right|$, $\left| \mathcal{E} \right|$, $\left| F \right|$, and $\left| C \right|$ denote the number of nodes, edges, features, and classes, respectively. Avg.D denotes the average degree of test nodes. EHR denotes the edge homophily ratio.}
\label{table:dataset}
\end{table}

\subsection{Implementation of Perturbations}
%\noindent
%{\bf \large Implementation of Perturbations.}
We examine the generalization of our model under three scenarios of perturbations, whose settings follow \cite{zhuang2022robust}. \\
1) Random perturbations ({\bf rdmPert}) simulate a non-malicious perturbator in OSNs that randomly connects with many other normal users for commercial purposes. In the experiments, we denote the percentage of random-perturbator nodes over validation/test nodes (a.k.a. victim nodes) as $p_{rdm}$ and restrict the number of connections from each perturbator to $\frac{1}{p_{rdm}}$. Except for specific statements, we select $p_{rdm} = 1\%$ in this paper. Note that rdmPert doesn't apply gradient-based attacks, such as FGSM \cite{goodfellow2014explaining}, PGD \cite{madry2017towards}, etc. \\
2) Information sparsity ({\bf infoSparse}) sparsifies 90\% links and 100\% features of the victim nodes ($L\&F$) on validation/test graphs. \\
3) Adversarial attacks ({\bf advAttack}) execute node-level direct evasion non-targeted attacks~\cite{zugner2018adversarial} on both links and features ($L\&F$) of the victim nodes on sparse graphs as such kind of attacks may fail on denser graphs. To ensure the sparsity, we sparse the denser graphs (AMZcobuy and Coauthor) and select the victim nodes whose total degrees are within (0, 10) for attacks. The intensity of perturbation $n_{pert}$ is set as 2 for all datasets. The ratio of $n_{pert}$ between applying on links and applying on features is 1: 10.

\subsection{Hyper-parameters and Settings of Our Model}
%\noindent
%{\bf \large Hyper-parameters of Our Model.}
Our experiments involve four two-layer classic GNNs (GCN~\cite{kipf2016semi}, GraphSAGE~\cite{hamilton2017inductive}, SGC~\cite{wu2019simplifying}, and GAT~\cite{velivckovic2018graph}) with 200 hidden units and a ReLU activation function. Their hyper-parameters are summarized in Tab.~\ref{tab:gnn_archit}. Within these GNNs, we select GCN as the backbone node aggregator for our model. Both these four GNNs and our model are trained by the Adam optimizer with a $1 \times 10^{-3}$ learning rate and converged within 200 epochs of training on all datasets. The weights of these node aggregators are initialized by Glorot initialization. Only the weight $\mathbf{W}_z$ in our model is initialized by He initialization. The size of weights is summarized in Tab.~\ref{tab:weights}. We set all $\lambda$ values as 1.0 in the experiments. We select the $\gamma_{min}$ based on the best validation accuracy. For retraining, our model is retrained to 500 epochs in total. In the ablation study, generalization test, and comparison of competing methods (including the inference runtime), we run experiments five times and report the mean values (some cases are reported as ``mean (std)"). For convergence verification, analysis of parameters, and further explorations, we only run the experiment once.

\begin{table}[h]
\scriptsize
%\footnotesize
%\small
\centering
\setlength{\tabcolsep}{2pt}
  \begin{tabular}{cccccc}
    \toprule
    \textbf{Model} & \textbf{\#Hops} & \textbf{Aggregator} & \textbf{\#Heads} & \textbf{Activation} & \textbf{Dropout} \\
    \midrule
    $\mathbf{GCN}$ & $\times$ & $\times$ & $\times$ & ReLU & 0.5 \\
    $\mathbf{SGC}$ & 2 & $\times$ & $\times$ & $\times$ & 0.5 \\
    $\mathbf{GraphSAGE}$ & $\times$ & gcn & $\times$ & ReLU & 0.5 \\
    $\mathbf{GAT}$ & $\times$ & $\times$ & 3 & ReLU & 0.5 \\
  \bottomrule
\end{tabular}
\caption{Model architecture of four classic GNNs.}
\label{tab:gnn_archit}
\end{table} % $\checkmark$

\begin{table}[h]
%\small
\centering
\setlength{\tabcolsep}{5pt}
  \begin{tabular}{c|c|c|c|c|c}
    \toprule
    $\mathbf{Weights}$ & $\mathbf{W}_{h}^{(0)}$ & $\mathbf{W}_{\mu}$ & $\mathbf{W}_{\sigma}$ & $\mathbf{W}_{z}$ & $\mathbf{W}_{h}^{(1)}$ \\
    \midrule
    $\mathbf{Size}$ & $d \times h$ & $d \times h$ & $d \times h$ & $ |\mathbf{Z}| \times h$ & $h \times K$ \\
    %$\mathbf{Layers}$ & $\mathbf{H}^{(1)}$ & $\mu$ & $log\sigma$ & $\mathbf{H}^{(1)}$ & $\mathbf{H}^{(2)}$ \\
  \bottomrule
\end{tabular}
\caption{Size of weights in our model. d, h, $|\mathbf{Z}|$, and K denote the number of features, hidden units, nodes in $\mathbf{Z}$, and classes, respectively.}
\label{tab:weights}
\end{table}

\begin{table}[t]
\scriptsize
%\footnotesize
%\small
\centering
\setlength{\tabcolsep}{2pt}
\begin{tabular}{ccccccc}
 \toprule
  {\bf Hyper-parameters} & {\bf Cora} & {\bf Citeseer} & {\bf Pubmed} & {\bf AMZcobuy} & {\bf Coauthor} & {\bf Flickr} \\
 \midrule
    Hidden units & 200 & 200 & 200 & 200 & 200 & 200 \\
    Dropout rate & 0.8 & 0.8 & 0.5 & 0.5 & 0.5 & 0.3 \\
    Learning rate & 1e-2 & 9e-3 & 1e-2 & 1e-2 & 1e-2 & 1e-2 \\
    Weight decay & 5e-3 & 1e-3 & 1e-3 & 1e-2 & 1e-3 & 1e-3 \\
    Use BN & $\times$ & $\times$ & $\times$ & $\checkmark$ & $\times$ & $\times$ \\
 \bottomrule
\end{tabular}
\caption{Hyper-parameters of DropEdge in this paper.}
\label{tab:dropedge_para}
\end{table}

\begin{table}[t]
\scriptsize
%\footnotesize
\centering
\setlength{\tabcolsep}{1.7pt}
\begin{tabular}{ccccccc}
 \toprule
  {\bf Hyper-parameters} & {\bf Cora} & {\bf Citeseer} & {\bf Pubmed} & {\bf AMZcobuy} & {\bf Coauthor} & {\bf Flickr} \\
 \midrule
    Propagation step & 8 & 2 & 5 & 5 & 5 & 5 \\
    Data augmentation times & 4 & 2 & 4 & 3 & 3 & 4 \\
    CR loss coefficient & 1.0 & 0.7 & 1.0 & 0.9 & 0.9 & 1.0 \\
    Sharpening temperature & 0.5 & 0.3 & 0.2 & 0.4 & 0.4 & 0.3 \\ 
    Learning rate & 0.01 & 0.01 & 0.2 & 0.2 & 0.2 & 0.2 \\
    Early stopping patience & 200 & 200 & 100 & 100 & 100 & 200 \\
    Input dropout & 0.5 & 0.0 & 0.6 & 0.6 & 0.6 & 0.6 \\
    Hidden dropout & 0.5 & 0.2 & 0.8 & 0.5 & 0.5 & 0.5 \\
    Use BN & $\times$ & $\times$ & $\checkmark$ & $\checkmark$ & $\checkmark$ & $\checkmark$ \\
 \bottomrule
\end{tabular}
\caption{Hyper-parameters of GRAND in this paper.}
\label{tab:grand_para}
\end{table}

\subsection{Hyper-parameters of Competing Methods}
%\noindent
%{\bf \large Hyper-parameters of Competing Methods.}
We compare our model with nine competing methods. GCN-Jaccard, GCN-SVD, and DropEdge belong to pre-processing methods. GRAND, RGCN, and ProGNN are categorized as inter-processing methods. GraphSS and LInDT are classified as post-processing methods. Both MC Dropout and post-processing methods are used for measuring the uncertainty purposes.
We present hyper-parameters of competing methods for reproducibility purposes. All models are trained by Adam optimizer. Note that GCN-Jaccard, GCN-SVD, DropEdge, ProGNN, MC Dropout, GraphSS, and LInDT also select GCN as the backbone model.
\begin{itemize}
\item {\bf GCN-Jaccard}~\cite{wu2019adversarial} preprocesses the graph by eliminating suspicious connections, whose Jaccard similarity of node’s features is smaller than a given threshold. The similarity threshold is 0.01. Hidden units are 200. The dropout rate is 0.5. Training epochs are 300.
\item {\bf GCN-SVD}~\cite{entezari2020all} proposes another preprocessing approach with low-rank approximation on the perturbed graph to mitigate the negative effects from high-rank attacks, such as Nettack~\citep{zugner2018adversarial}. The number of singular values is 15. Hidden units are 200. The dropout rate is 0.5. Training epochs are 300.
\item {\bf DropEdge}~\cite{rong2019dropedge} randomly removes several edges from the input graph in each training epoch. We choose 1 base block layer and train this model with 300 epochs. Other parameters are mentioned in Tab.~\ref{tab:dropedge_para}.
\item {\bf GRAND}~\cite{feng2020graph} proposes random propagation and consistency regularization strategies to address the issues of over-smoothing and non-robustness of GCNs. The model is trained with 200 epochs. The hidden units, drop node rate, and L2 weight decay are 200, 0.5, and $5 \times 10^{-4}$. Other parameters are reported in Tab.~\ref{tab:grand_para}.
\item {\bf RGCN}~\cite{zhu2019robust} adopts Gaussian distributions as hidden representations of nodes to mitigate the negative effects of adversarial attacks. We set up $\gamma$ as 1, $\beta_1$ and $\beta_2$ as $5 \times 10^{-4}$ on all datasets. The hidden units for each dataset are 64, 64, 128, 1024, 1024, and 2014, respectively. The dropout rate is 0.6. The learning rate is 0.01. The training epochs are 400.
\item {\bf ProGNN}~\cite{jin2020graph} jointly learns the structural graph properties and iteratively reconstructs the clean graph to reduce the effects of adversarial structure. We select $\alpha$, $\beta$, $\gamma$, and $\lambda$ as $5 \times 10^{-4}$, 1.5, 1.0, and 0.0, respectively. The hidden units are 200. The dropout rate is 0.5. The learning rate is 0.01. Weight decay is $5 \times 10^{-4}$. The training epochs are 100.
\item {\bf MC Dropout}~\cite{gal2016dropout} develops a dropout framework that approximates Bayesian inference in deep Gaussian processes. We follow the same settings as our proposed model except for selecting the optimal dropout rate for each dataset as 0.7, 0.3, 0.9, 0.6, 0.8, and 0.3, respectively. Note that MC Dropout will apply dropout in test data.
\item {\bf GraphSS}~\cite{zhuang2022defending} approximates inferred labels to original labels on perturbed graphs as closely as possible by iterative Bayesian label transition. We follow the same settings as GraphSS. The number of epochs for inference, retraining, and warm-up steps is 100, 60, and 40, respectively. The $\alpha$ is fixed as 1.0.
\item {\bf LInDT}~\cite{zhuang2022robust} proposes a new topology-based label sampling method with dynamic Dirichlet prior to boost the performance of Bayesian label transition against topological perturbations. We follow the same settings as LInDT. The number of inference epochs is 100, 200, 80, 100, 90, and 100 for each dataset. The number of retraining epochs and warm-up steps remains at 60 and 40, respectively. The $\alpha$ value is 0.1, 0.3, 1.0, 0.7, 0.1, and 0.1 for each dataset in the beginning and will be dynamically updated during the inference.
\end{itemize}

\subsection{Evaluation Metrics}
%\noindent
%{\bf \large Evaluation Metrics.}
In this work, we evaluate the performance by classification accuracy (Acc.) and average normalized entropy (Ent.), which is formulated in Eq.~\ref{eqn:entropy}. Acc. is a classic metric to evaluate the performance of node classifications. Higher Acc. indicates better node classification. We use Ent. to evaluate the uncertainty. Lower Ent. portrays lower uncertainty on the nodes' categorical distributions.
\begin{equation}
\mathbf{Ent}(\mathbf{X}) = -\frac{1}{N} \sum_{i=1}^{N} \sum_{j=1}^{d} \frac{p(x_{i,j})\ln{(p(x_{i,j}))}}{\ln{(d)}}.
\label{eqn:entropy}
\end{equation}

\subsection{Hardware and Software}
%\noindent
%{\bf \large Hardware and Software.}
All experiments are conducted on the server with the following configurations:
\begin{itemize}
  \item Operating System: Ubuntu 18.04.5 LTS
  \item CPU: Intel(R) Xeon(R) Gold 6258R CPU @ 2.70 GHz
  \item GPU: NVIDIA Tesla V100 PCIe 16GB
  \item Software: Python 3.8, PyTorch 1.7.
\end{itemize}

\section{Supplementary Experiments}

\subsection{Comparison of Inference Runtime}
%\noindent
%{\bf \large Comparison of Inference Runtime.}
Tab.~\ref{table:avgtime} presents the average inference runtime among MCMC methods and our model in ten streaming sub-graphs. Our model has a significantly lower runtime over all datasets as our retraining mechanism only retrains the model once for the same type of perturbations, whereas MCMC methods must run the inference model on top of GNNs to output the inferred labels for each sub-graph. This advantage implies that our model can be applied to various online downstream tasks for robust inference.

\begin{table}[h]
%\scriptsize
\footnotesize
%\small
\centering
\setlength{\tabcolsep}{1pt}
\begin{tabular}{ccccccc}
  \toprule
      & {\bf Cora} & {\bf Citeseer} & {\bf Pubmed} & {\bf AMZcobuy} & {\bf Coauthor} & {\bf Flickr} \\
    \midrule
    \textbf{GraphSS} & 23.66 & 24.71 & 113.59 & 51.33 & 109.82 & 276.50 \\
    \textbf{LInDT} & 32.04 & 66.50 & 143.88 & 76.10 & 156.08 & 318.33 \\
    \textbf{Ours} & {\bf 2.02} & {\bf 3.29} & {\bf 10.78} & {\bf 4.19} & {\bf 9.31} & {\bf 21.35} \\
  \bottomrule
\end{tabular}
\caption{The average inference runtime (in seconds) among MCMC methods and our model in streaming settings.}
\label{table:avgtime}
\end{table}

\subsection{Discussion on Other Competing Methods}
%\noindent
%{\bf \large Discussion on Other Competing Methods.} 
Besides comparing the inference runtime, we have several observations in other competing methods, as follows: First, both post-processing methods and MC Dropout can help decrease the Ent. on all datasets. Second, RGCN obtains higher accuracy on larger graphs. We argue that adopting Gaussian distributions in hidden layers can reduce the negative impacts on larger perturbed graphs. Third, ProGNN performs better on denser graphs. This is possible because ProGNN utilizes the properties of the graph structure to mitigate perturbations while denser graphs preserve the plentiful topological relationships. From the perspective of runtime (in seconds), we observe that GCN-Jaccard, GCN-SVD, DropEdge, and MC dropout consume much less time than other methods. These are reasonable because these methods mainly process the graph structure by dropping edges or hidden units. However, these methods may not be effective on real-world social graphs as the graph structure of OSNs is dynamically changing. On the contrary, ProGNN consumes much more time as the size of graphs increases because it preserves low rank and sparsity properties via iterative reconstructions.

\begin{figure}[t]  % EX5
  %\hfill
  \begin{subfigure}{0.48\linewidth}
    \centering % includes the first image
    \includegraphics[width=\textwidth]{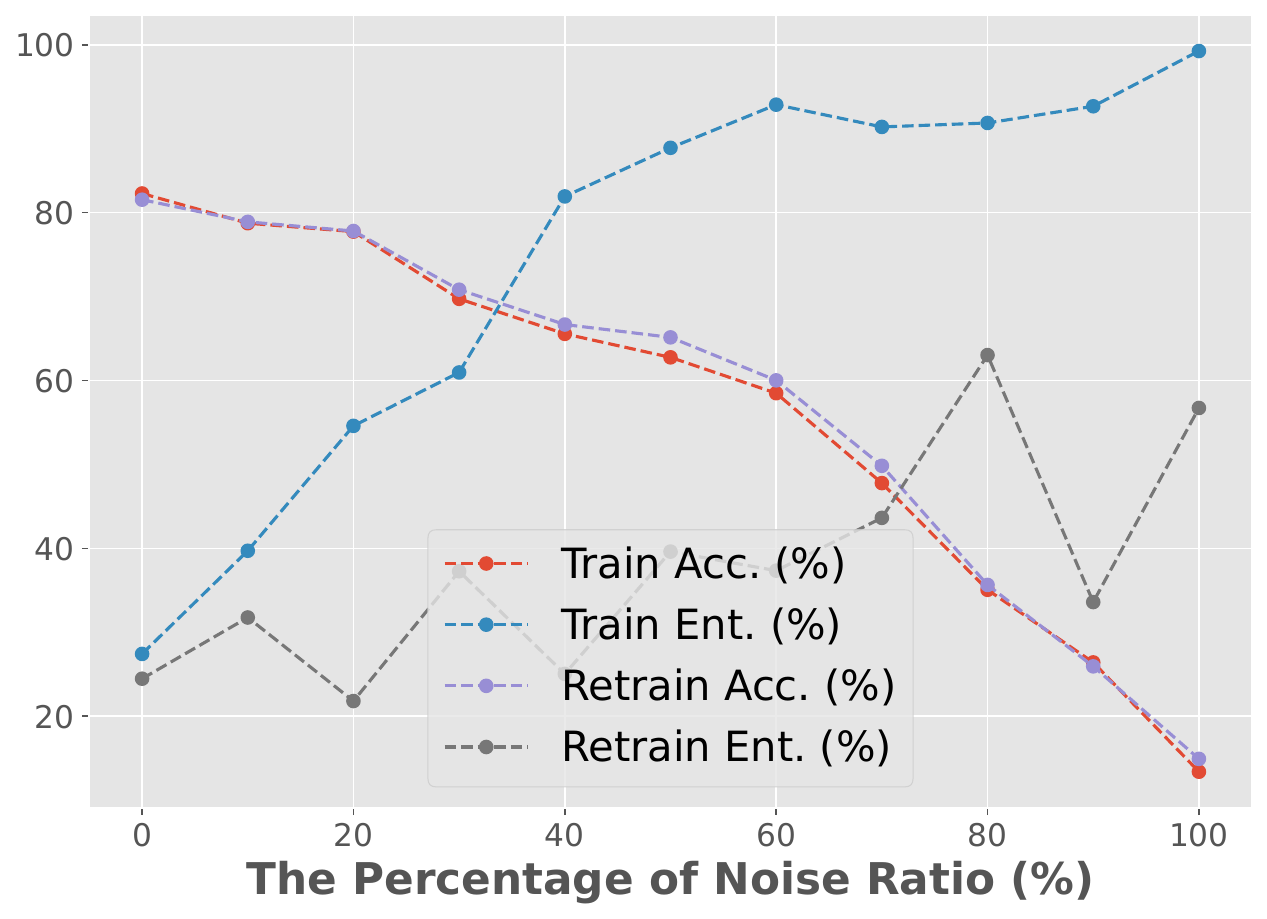}
  \end{subfigure}%
  %\hfill
  \begin{subfigure}{0.52\linewidth}
    \centering % includes the second image
    \includegraphics[width=\textwidth]{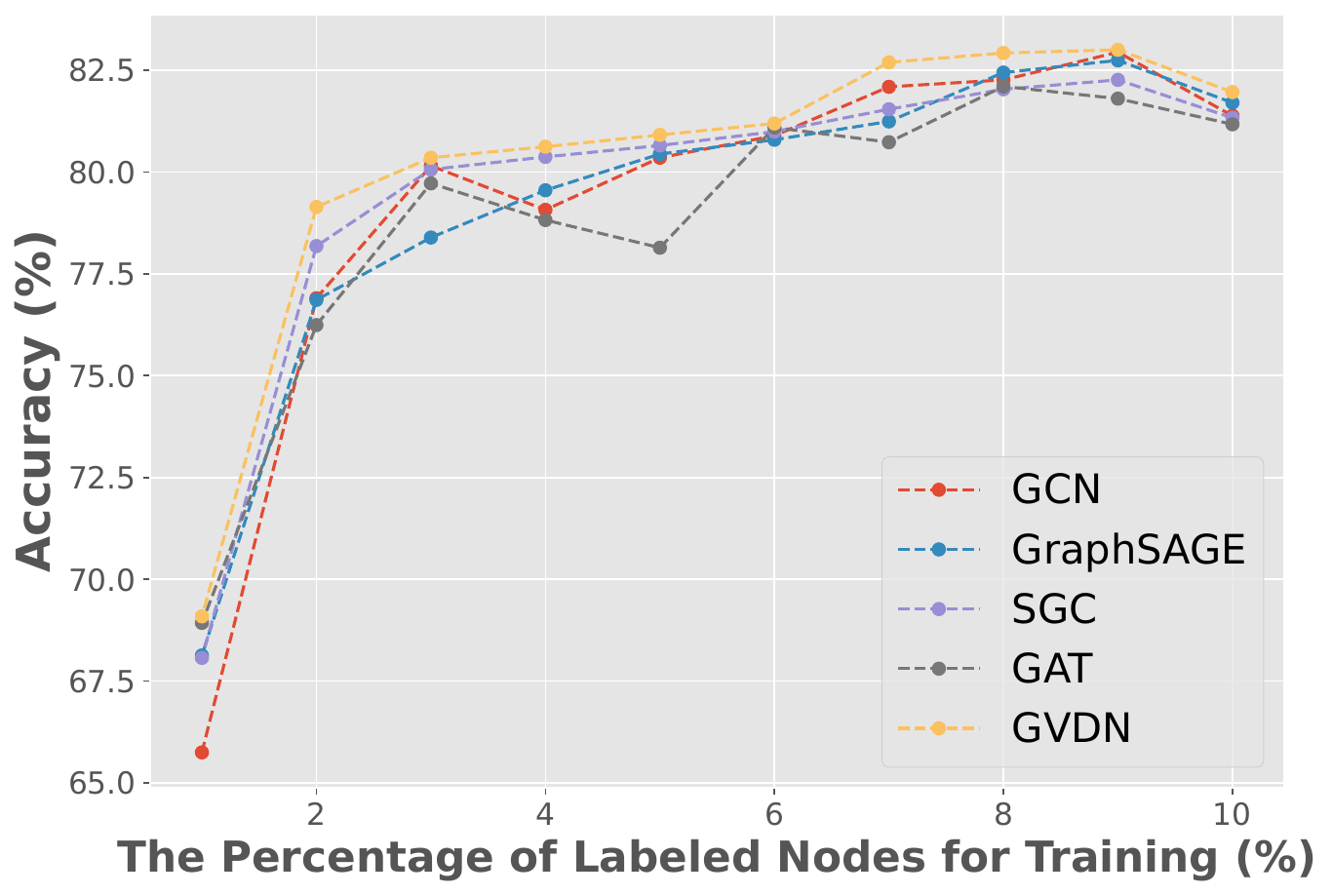}
  \end{subfigure}
\caption{Further explorations under the setting of noise labeling (left side) and sparse labeling (right side).}
\label{fig:fig_nr_lr}
\end{figure}

\subsection{Further Explorations}
%\noindent
%{\bf \large Further Explorations.}
Besides recovering the performance on the perturbed graphs, we explore whether our model can benefit other research fields on Cora.
We first investigate the performance of our model under the noisy labels setting (left side of Fig.~\ref{fig:fig_nr_lr}). In this experiment, we inject the noises into the train labels by randomly replacing the ground-truth labels with another label, chosen uniformly. We denote the percentage of the noises as noise ratio, $nr$, and adjust this ratio from 0\% to 100\%. As expected, the accuracy decreases as the $nr$ increases. However, our proposed retraining mechanism can maintain the normalized entropy at a relatively low level compared to the result before retraining. This result indicates that our model could contribute to the uncertainty estimation on the graphs with noisy labeling.
Furthermore, we examine whether our model can work under the sparse labeling setting (right side of Fig.~\ref{fig:fig_nr_lr}). In this setting, the percentage of the labeled train nodes changes from 1\% to 10\%. The results reveal that our model can maintain robust performance when the labeling rate is not less than 2\% and slightly outperforms the other four popular GNNs without retraining.

\end{document}